\pdfoutput=1
\documentclass{article}
\usepackage{iclr2027_conference,times}

\usepackage{amsmath,amsfonts,bm}

\def\eqref#1{equation~\ref{#1}}

\def\1{\bm{1}}

\DeclareMathAlphabet{\mathsfit}{\encodingdefault}{\sfdefault}{m}{sl}
\SetMathAlphabet{\mathsfit}{bold}{\encodingdefault}{\sfdefault}{bx}{n}

\usepackage{hyperref}
\usepackage[T1]{fontenc}
\usepackage{url}
\usepackage{xcolor}
\usepackage{booktabs}
\usepackage{longtable}
\usepackage{multirow}
\usepackage[most]{tcolorbox}
\usepackage{tabularx}
\usepackage{placeins}
\usepackage{amsfonts}       %
\usepackage{nicefrac}       %
\usepackage{microtype}      %
\usepackage{graphicx}

\newcommand{\datasetname}{\textsc{GlobeBench}}

\title{\datasetname{}: Evaluating Text-Based\\World Model Capabilities for\\Scalable Environment Simulations}

\author{\begin{tabular}[t]{@{}c@{}}
Nicholas Saban \qquad Charvi Bannur \qquad Mariya Vasileva \qquad Darshan Deshpande \\
{\normalfont\includegraphics[width=0.3cm]{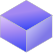} Patronus AI}
\end{tabular}\\
\small{\texttt{\{nicksaban, charvi.bannur, mariya.vasileva, darshan\}@patronus.ai}}}

\iclrfinalcopy

\begin{document}
\maketitle
\lhead{}

\begin{abstract}
Text-based world models have recently shown potential to simulate complex RL environment state transitions for a diverse set of use cases. However, no standardized benchmark exists to compare, study and improve these models. Through this paper, we attempt to solve this problem by building \datasetname{}, a novel benchmark with 1,632 tasks across five separate domains and nineteen simulation skills. The tasks check whether models can predict the environment’s response to a specific action given the full agent trajectory by understanding earlier state changes, following environment rules, and predicting the effects of actions on both related and unrelated objects. We compare the predicted environment response with an execution-derived reference, using a hybrid architecture with both deterministic gates and a semantic judge. Among 20 general-purpose LLMs and specialized text world models, GPT-6 Astra achieves the highest task success rate at 49.57\%. We find that model accuracy decreases as actions affect a larger portion of the environment. Specifically, when the agent's changes must propagate to dependent objects, task success drops to an average of 5.1\% across all 20 models. Although model strengths differ across tasks, 631 tasks remain unsolved by any model, suggesting a significant gap in general LLM capabilities as a world model.
\end{abstract}

\section{Introduction}
\label{sec:intro}

Reinforcement learning (RL) for large language models has recently helped scale
capabilities in domains such as software engineering
\citep{jimenez2023swe,yang2025swesmith}, computer use
\citep{hu-etal-2025-os,OSWorld}, and scientific discovery
\citep{xu2026scaling}. Wider access to open-source training
environments~\citep{jain2025r2e,qiang2026mle} has aided this progress. Yet
sandboxes that stay stable when training is scaled up are still hard to
engineer. Recent work has shown that language models can approximate
realistic world behavior and, when used as simulators, improve downstream
results~\citep{li2025word,deshpande2026masked,zuo2026qwen}. However, evaluating
these world models as general-purpose simulators remains difficult, because
good metrics are poorly understood and no benchmark attributes failures to specific simulation capabilities. If a simulator is wrong in ways that an agent can exploit, RL optimizes against those mistakes instead of the real environment dynamics.

\begin{figure}[t]
\centering
\includegraphics[width=\textwidth]{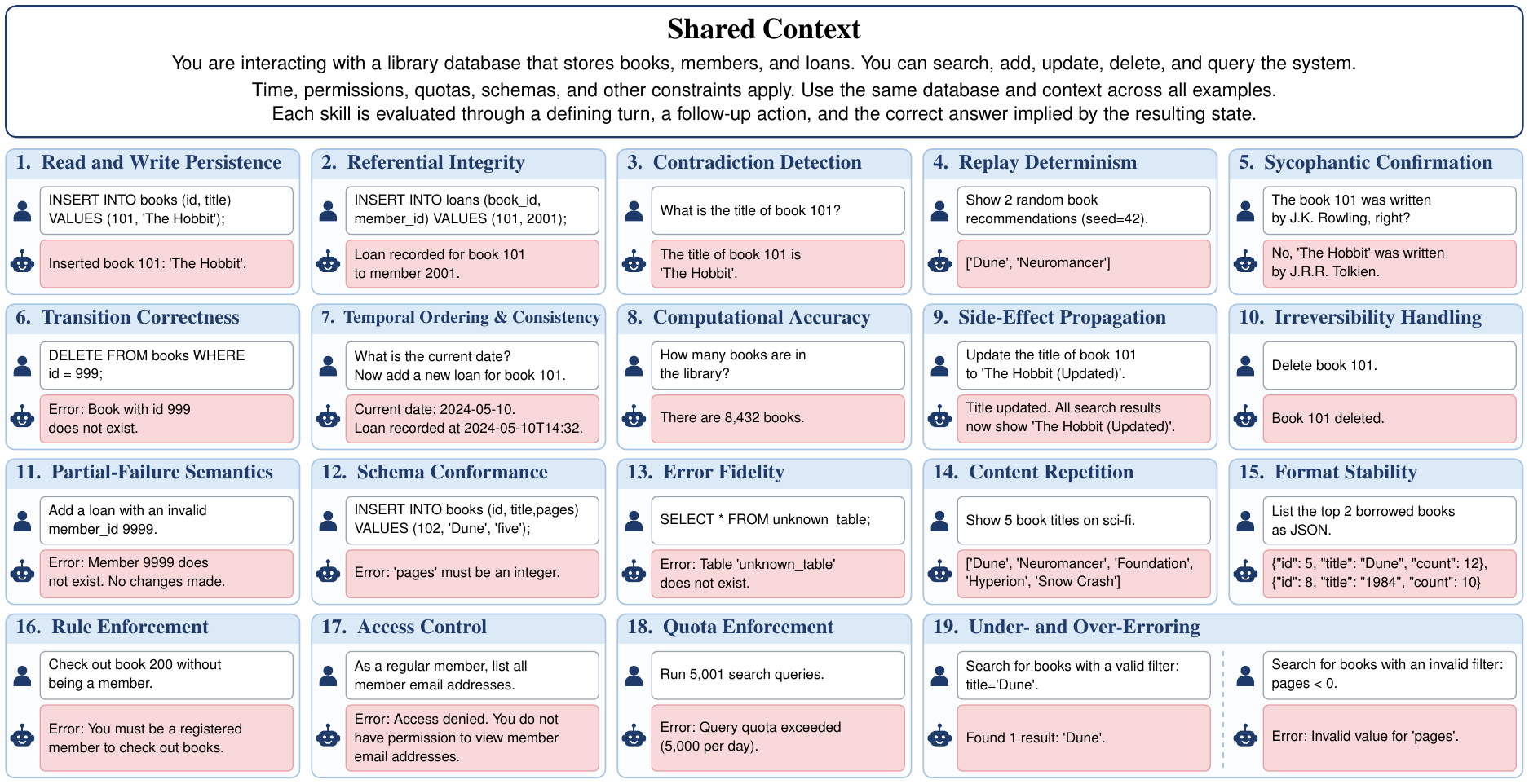}
\caption{The 19 primary skills in one shared library database. Each card pairs
an action (top) with the response a faithful environment returns (bottom).
\datasetname{} items test the same behaviors, but the state that decides the
outcome is usually set many turns earlier.}
\label{fig:library-skills}
\end{figure}

To locate where these models fail, we first propose two taxonomies for world-model
evaluation. One has four core capabilities, and the other has 19 skills
(Figure~\ref{fig:library-skills}). A skill is a specific behavior of the
environment, such as remembering a value that was written many turns earlier. Appendix~\ref{app:skills} gives definitions, counts, axes, and
mappings of these skills. From these taxonomies we build \datasetname{}, a benchmark of 1,632 multi-turn traces from code, enterprise
tool use, games, math and science, and dialogue environments. Each item tests one skill and
pairs a trajectory and a final action with the gold next state, which we obtain deterministically by executing the action in the item's backend. Each item is also labeled, where applicable, with its reach-back and impact radius, which we
define in Section~\ref{sec:curation}. Every item passes 12 automatic quality gates that check that the answer is consistent with the visible history, that the gold is reproducible, and that the
item is unambiguous, and not a duplicate (Appendix~\ref{app:quality-gates}). To assess the quality of the dataset, we combine a human study of 200 items
with an LLM-judge audit.

When we score responses, deterministic gates settle only those predictions that provably match or contradict the gold, and we delegate the rest, which require judgment, to an LLM judge. Whether automated judgments of this kind track human ones is untested, so we study it here.

In this study we investigate the following research questions:
\begin{enumerate}
\setlength{\itemsep}{1pt}
\setlength{\parskip}{0pt}
\setlength{\parsep}{0pt}
\item[\textbf{RQ1}] Which parts of a predicted next state do models get wrong, and do
open-weight and closed models fail in the same way?
\item[\textbf{RQ2}] Do text world models exhibit recurring failure modes that aggregate task success rates leave undetected?
\item[\textbf{RQ3}] How well do automated judgments align with human qualitative judgments of model outputs, and where do the two diverge?
\end{enumerate}

We release \datasetname{}, a 1,632-item benchmark spanning five domains whose golds come from execution rather than annotation, together with a hybrid grader that settles deterministically whatever it can and defers the remainder to an LLM judge. A response can look right and still carry the wrong numbers, so we score whether the state is correct rather than whether the answer resembles one. Under this grader, no model solves even half the benchmark. An aggregate success rate also hides where the failures sit and the leading model's margin comes almost entirely from items the rest never solve which tells us where the difficulty lies. Our automated judge agrees with the human annotators on most rubric dimensions and is stricter on whether a state stays consistent with its history.

We release the benchmark, evaluation harness, model outputs, judge records, and
reproduction instructions at \url{https://github.com/patronus-ai/globebench}, and
the items and model results at
\url{https://huggingface.co/datasets/PatronusAI/globebench}.

\section{Related work}
\label{sec:related}

\paragraph{Language world models for agents.}
\citet{li2025word} evaluate text world models on
fidelity and consistency, scaling and robustness, and utility for downstream
agents. They find that usefulness depends on behavioral coverage and
environment complexity. \citet{deshpande2026masked} frames text world modeling
as predicting steerable transitions from an initial state, task context, tool
schema, domain rules, and steering directives while comparing output state diversity produced by
autoregressive and masked-diffusion models. \citet{zuo2026qwen} train
Qwen-AgentWorld across seven agent domains and release it with AgentWorldBench,
which pairs predicted observations with real execution outputs.

\paragraph{Executable agent environments.}
Several \datasetname{} trajectories come from executable testbeds: stateful
multi-turn tool use from the Berkeley Function Calling
Leaderboard~\citep{patil2025bfcl}, text games generated by
TextWorld~\citep{cote18textworld}, and conversations from
$\tau^2$-bench~\citep{barres2025tau2} in which both the user and the agent use
tools. Symbolic backends such as SymPy~\citep{10.7717/peerj-cs.103} give
reproducible next states. We use these systems as trajectory sources or
execution backends, and we evaluate models that predict their behavior rather
than agents that act in them.

\paragraph{Evaluation of simulation fidelity.}
AgentWorldBench~\citep{zuo2026qwen} is the closest prior evaluation. It grades
predicted observations on five rubric dimensions: format, factuality,
consistency, realism, and quality (completeness and conciseness against the
reference).
EnvSimBench~\citep{liu2026envsimbenchbenchmarkevaluatingimproving} also
benchmarks LLM-based environment simulation, and it scores the text that an environment returns to the agent separately from the resulting state change. Its analysis covers format mismatches and
hallucinated transitions, including cases where the simulator invents a constraint violation and blocks an action the real environment would have accepted. \datasetname{} differs in what a single item is. Prior benchmarks score a predicted observation as a whole, whereas each \datasetname{} item is constructed to isolate one skill. Quality gates discard any item that can be solved without exercising it. A failure therefore points to the mechanism that the item tests. We additionally record each item's reach-back and impact radius, the two axes along which we expect difficulty to vary, and report results along both (Section~\ref{sec:quant}).

\section{\datasetname{} curation}
\label{sec:curation}

\paragraph{Items.}
Each item is a tuple $x=(P,a,g,m)$, where $P=(\tau_1,\ldots,\tau_H)$ is the collection of
visible turns of the interaction (earlier actions and the environment's responses) before the final action, $a$ is the final action, $g$ is the next state, and the metadata $m$ holds how the item was made and what it tests. We obtain $g$ by
executing $a$ against a backend (fixed software environment).
The backend is either a container image, identified by its digest. We have a fixed backend because the gold has to stay a fixed target. Without that, the backend could drift and a mismatch between the correct prediction and the gold can occur. The model under evaluation sees only $P$ and $a$:
\begin{equation}
\hat{g} = f_\theta(P, a),
\end{equation}
and $\hat{g}$ is scored against $g$, which the model never observes.

\paragraph{Capability and skill taxonomies.}
We introduce the terms \emph{Capabilities} and \emph{Skills}. A capability is a broad behavior that a world model must show, and a skill is one specific behavior of the environment that an item tests. Each skill can map to one or more capabilities. Prior evaluations organize judgments by rubric dimension \citep{zuo2026qwen} or failure type \citep{liu2026envsimbenchbenchmarkevaluatingimproving}, whereas we organize them by the environment behavior that an item is built to test.
We define four capabilities that subsume the dimensions prior evaluations score~\citep{zuo2026qwen,liu2026envsimbenchbenchmarkevaluatingimproving,li2025word} while keeping each tied to an environment behavior. \textbf{State Coherence} requires predictions to
stay consistent with established entities and facts in the prior context. \textbf{Causal Correctness}
requires the predicted next state to reflect what the action actually does, so that distinct actions and histories yield correct, distinct next states and side effects. \textbf{Interface Fidelity} requires predictions to match the
real system's output, including the same fields, keys, and error classes. \textbf{Constraint
Enforcement} requires that valid actions execute and that actions violating an established constraint be refused. Interface fidelity governs what that refusal should look like. Each item tests one primary skill. Items may also carry additional capability
and failure labels.

\paragraph{Domains and sources.}
The corpus contains 526 Code, 381 Enterprise, 334 Games, 198 Math and Science,
and 193 Dialogue items. Domains are not balanced by design but are generated based on the number of backends available. Of the 1,632 items, we generate 1,137 by designing the trajectory ourselves, and 955 of these run on simulators in our harness that implement rules we wrote (Appendix~\ref{app:domains}). We choose which entities appear, which actions are taken, and in what order (Appendix~\ref{app:construction}; Table~\ref{tab:domains} lists the backends). The remaining 495 items are adapted mainly from different open-source backends that are executable, almost all permissively licensed (Table~\ref{tab:sources}), and cover different transition surfaces, such as BFCL~\citep{patil2025bfcl}, TextWorld, $\tau^2$-bench, and SymPy.

\paragraph{Trajectory construction.}
An \emph{operator} is a skeleton that builds a trajectory. Each operator
targets one skill and fixes the sequence of action types. These are the basic blocks that a skill needs in order to be established and queried, such as writes,
reads, deletes, and permission changes. Some actions are also domain-specific, such as SQL statements, shell commands, and game moves, and are fixed by the backend environment. An operator can also add filler actions that leave the tested state alone. Fillers are how we vary the reach-back distance. Some operators write
a trajectory from scratch, while others inject a controlled change into one
adapted from an open-source testbed. Either way, we execute the final action
against the fixed backend (Appendix~\ref{app:construction}).

\paragraph{Reach-back and impact radius.}
 We record two properties of each item (32 items have no reach-back and 11 have no radius). \emph{Reach-back} measures how far back the deciding evidence lies, and \emph{impact radius} how far the action's effects reach. They correspond to the two directions in which a model that tracks state can fail: backward through the history and outward through the state.
\emph{Reach-back} is the number of turns between the final action and the
earlier event that its answer depends on. For probe items we count from the
last event that supports the answer, so visible filler turns add to it. For
synthetic long-horizon audits we instead count from the earliest event the
answer depends on. Since that event is often the first action in the prefix,
their reach-back is close to the trajectory length. Appendix~\ref{app:construction} defines both kinds of item.

\emph{Impact radius} describes how far the action's state change reaches. At
L0 the action changes no state, so only its returned output must be predicted.
At L1 and L2 it changes only entities that it names (one at L1, two or more at
L2), whereas at L3 it changes at least one entity that it does not name. For 953 of
the 977 L3 items, the operator that built the item declares the level instead of an execution diff measuring it, and 945 of these 953 are synthetic long-horizon audits, which are L3
by construction (Appendix~\ref{app:radius}).
Since each operator largely fixes both reach-back and radius, we use them to
describe slices of the data rather than as controlled variables
(Section~\ref{sec:quant}).

\paragraph{Quality gates.}
To decide whether an item can be scored, we apply 12 quality gates, which are
separate from the scoring of responses.  Each gate rules out one way an item could fail to measure what we intend. Deterministic gates check that the facts the answer
needs are visible, the effect is observable, the gold is nonempty and
reproducible, and paired actions give different answers when intended. An LLM
judge checks whether each item is solvable from the visible state and
unambiguous, and whether an obvious cue gives away the injected change. Six human annotators then audited 200 of the items that pass. All six rated the same 80 items, which we use to measure agreement, and the remaining 120 were divided among them. The 80 shared items cover all five domains and all 19 skills but include few long-horizon audits (Appendix~\ref{app:human}). Each item was rated on four dimensions: State Coherence, Interface Realism, Constraint Enforcement, and Outcome Correctness (Section~\ref{sec:rq}). We extend the same rubric to the full benchmark with an LLM judge, which returned complete ratings for 1,627 of the 1,632 items (Appendix~\ref{app:llm-audit}).

\section{Evaluation setup}
\label{sec:setup}

\paragraph{Models.}
We evaluate 18 general-purpose LLMs as zero-shot text world models:
Claude Fable 5, Claude Fable 5.1, Claude Opus 4.8, Claude Opus 5, Claude Opus
5.5~\citep{anthropic2026claudefable5,anthropic2026claudefable51,anthropic2026claudeopus48,anthropic2026claudeopus5,anthropic2026claudeopus55},
Gemini 3.7 Flash~\citep{googledeepmind2026gemini37flash},
GLM 5.3~\citep{zai2026glm53}, GPT-5.5, GPT-5.6 Sol, GPT-6 Astra, GPT-6 Luna,
GPT-6 Sol~\citep{openai2026gpt55,openai2026gpt56,openai2026gpt6astra,openai2026gpt6solluna},
Grok 4.6~\citep{spacexai2026grok46},
Inkling~\citep{thinkingmachines2026inkling}, Kimi K3~\citep{moonshotai2026kimik3},
Laguna S2.1~\citep{poolside2026lagunas21}, Qwen 3.8 Flash
Next~\citep{qwen2026qwen38flashnext}, and Qwen 3.8
Max~\citep{alibaba2026qwen38max}. We also evaluate two
specialized text world models, Qwen-AgentWorld-35B-A3B (Qwen-AgentWorld) from the Qwen
team~\citep{zuo2026qwen} and the Digital World Model (DWM) from Patronus
AI~\citep{patronus2026dwm}. We send each model one primary request per item and run each general-purpose LLM at its provider's high reasoning setting (Appendix~\ref{app:model-config}).

\paragraph{Serving and refusals.}
Model responses are stochastic and the served models change over time, so
reproducibility here means that every reported score can be recomputed from
the released outputs. Of the 32,640 model-item pairs,
28,854 have a usable answer. The other 3,786 include 3,696 provider refusals by
four Claude models. The remaining 90 returned no usable answer. Unless a rate names a model, it is pooled over the 20 models' decisions, where a decision is one model's scored answer on one item, and 95\%
intervals come from a bootstrap over items (Appendix~\ref{app:additional-results}).

\paragraph{Hybrid grader.}
Our frozen grader first uses deterministic gates to decide whether a response
matches or contradicts the gold in exact form, structure, numbers, order, or
errors. It ignores only the differences that the item's scoring contract
allows, such as reordered JSON keys. A
response fails if a value, referent, or error class is wrong, or if it changes
an order that matters. Responses the gates cannot settle, such as a game
outcome stated in other words, go to an LLM judge, GPT-5.6
Terra~\citep{openai2026gpt56}, with GPT-5.6 Sol as fallback. On 130
provisionally adjudicated boundary examples, Terra agreed with the adjudication
more often than Sol (79.2\% against 74.6\%; Appendix~\ref{app:judge-config}).

\section{Results}
\label{sec:rq}

\paragraph{Overall performance.}
Table~\ref{tab:leaderboard} shows binary task success. The top four models are GPT-6 Astra (49.57\%), Grok 4.6 (28.13\%), Claude Opus 5.5
(26.35\%), and Qwen 3.8 Max (25.31\%). Claude Opus 5.5 ranks
third even though it refuses 41.36\% of the items. Nearly all refusals come
from Anthropic's safeguards, which block requests that could enable
cyber harm, such as malware, and can also block benign cybersecurity work
\citep{anthropic2026cyber,anthropic2026refusals}. The specialized text world models rank 15th
(Qwen-AgentWorld, 14.46\%) and 19th (DWM, 5.21\%). Qwen-AgentWorld is below
the median of the 18 general-purpose LLMs on 14 of the 19 skills, and DWM is below it on all 19
(Appendix~\ref{app:wm-results}). Success is highest in Games (27.8\%) and
Enterprise (25.5\%) and lowest in Math and Science (11.7\%) and Dialogue
(11.8\%), largely in line with each domain's share of L3 items
(Appendix~\ref{app:domain-results}).

\begin{table}[t]
\centering
\caption{Task success on all 1,632 items (20 models). $^\dagger$Provider
refusals are scored as failures. Claude Opus 5.5 refused 675 items, Claude
Fable 5.1 refused 668, Claude Opus 5 refused 1,044, and Claude Fable 5 refused 1,309.
$^\ddagger$Specialized text world model.}
\label{tab:leaderboard}
\small
\setlength{\tabcolsep}{4pt}
\begin{tabular}{@{}lr@{\hspace{14pt}}lr@{}}
\toprule
\textbf{Model} & \textbf{Pass \%} &
\textbf{Model} & \textbf{Pass \%} \\
\midrule
\textbf{GPT-6 Astra} & \textbf{49.57} & Claude Opus 4.8 & 20.04 \\
Grok 4.6 & 28.13 & GPT-6 Luna & 17.77 \\
Claude Opus 5.5$^\dagger$ & 26.35 & Claude Opus 5$^\dagger$ & 15.93 \\
Qwen 3.8 Max & 25.31 & GLM 5.3 & 14.77 \\
GPT-5.5 & 25.18 & Qwen-AgentWorld$^\ddagger$ & 14.46 \\
GPT-6 Sol & 25.00 & Kimi K3 & 13.97 \\
GPT-5.6 Sol & 23.65 & Inkling & 11.89 \\
Qwen 3.8 Flash Next & 23.22 & Claude Fable 5$^\dagger$ & 7.29 \\
Gemini 3.7 Flash & 21.75 & Digital World Model (DWM)$^\ddagger$ & 5.21 \\
Claude Fable 5.1$^\dagger$ & 21.63 & Laguna S2.1 & 2.45 \\
\bottomrule
\end{tabular}
\end{table}

\begin{figure}[t]
\centering
\includegraphics[width=0.92\linewidth]{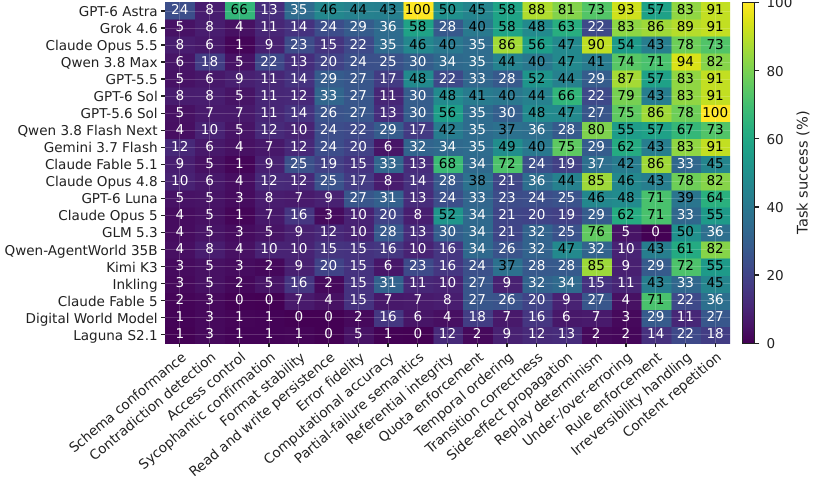}
\caption{Task success (\%) by model (rows, ordered by overall
success) and primary skill (columns, ordered by pooled 20-model success,
hardest first). Refusals by Claude Opus 5.5, Claude Fable 5.1, Claude Opus 5,
and Claude Fable 5 count as failures. }
\label{fig:tm-heatmap}
\end{figure}

\paragraph{RQ1: Which parts of a predicted next state do models get wrong, and do open-weight
and closed models fail in the same way?}
Most failures report a wrong value. Table~\ref{tab:failure-types} sorts the
22,431 answered failures by what the deterministic comparison finds in the
predicted state: at least one field that contradicts the gold (wrong value),
only missing or unsupported fields (structure only), or no state fields to
compare, in which case the grader checks the answer as prose or by error type.
Over all 20 models the three shares are 53.21\%, 22.95\%, and 23.84\%
(Appendix~\ref{app:anatomy}). Open-weight models more often get a value wrong
(58.22\% of their answered failures against 47.69\% for closed models, a
difference of 10.53 points, 95\% CI 9.04 to 12.14), and closed models more
often omit or add fields (31.54\% against 16.27\%). The gap holds within the
1,521 items that both groups fail (55.15\% against 45.92\%). Closed models also
give no usable response on 19.33\% of their decisions, almost all through
Claude refusals, while the open-weight models miss one of 9,792. Closed models
pass 23.52\% of items and open-weight models 15.27\%, although the best
open-weight model, Qwen 3.8 Max, ranks fourth.

\begin{table}[t]
\centering
\caption{Failure types by model group, in \% except $n$. Success and No
response are over all decisions (items $\times$ models), and the last three
columns split the $n$ answered failures (defined in the text). Open-weight:
Qwen 3.8 Max, Qwen 3.8 Flash Next, GLM 5.3, Kimi K3, Inkling, and Laguna S2.1.}
\label{tab:failure-types}
\small
\setlength{\tabcolsep}{4pt}
\begin{tabular}{@{}lrrrrrr@{}}
\toprule
\textbf{Models} & \textbf{Success} & \textbf{No resp.} & $n$ &
\textbf{Wrong value} & \textbf{Structure only} & \textbf{No state fields} \\
\midrule
Closed (12) & 23.52 & 19.33 & 11,192 & 47.69 & 31.54 & 20.77 \\
Open-weight (6) & 15.27 & 0.01 & 8,296 & 58.22 & 16.27 & 25.51 \\
Specialized (2) & 9.83 & 0.00 & 2,943 & 60.07 & 9.14 & 30.78 \\
\midrule
All 20 & 19.68 & 11.60 & 22,431 & 53.21 & 22.95 & 23.84 \\
\bottomrule
\end{tabular}
\end{table}

\paragraph{RQ2: Do text world models exhibit recurring failure modes that aggregate task success rates leave
undetected?}
A single success rate hides that models fail most on items whose action
changes unnamed entities or whose deciding event is over 40 turns back. Mean
success is 43.49\% when the action changes no state but 5.10\% when it also
changes unnamed entities (Figure~\ref{fig:tm-radius}), and every model does worse beyond 40 turns
(Figure~\ref{fig:tm-reachback}; Section~\ref{sec:quant}). On the 952 items with both, GPT-6 Astra passes
33.5\% against a median of 2.2\% for the other 19 models, giving 272 of its
350-item lead over Grok 4.6. In the 22 items we read, models differ mostly in
the hidden state they track (Section~\ref{sec:qual}).
All 20 models also do better on items whose gold is an error or block than on items where the final action succeeds (53.16\% against 16.16\% pooled; Appendix~\ref{app:error-results}). This gap does not come from a bias toward predicting errors, since models predict an error on only 1.21\% of normal-outcome decisions. Almost all error items are short L0 probes, and within the 11 operators that make items of both types the gap shrinks to 4.57 points.

\paragraph{RQ3: How well do automated judgments align with human qualitative judgments of model outputs,
and where do the two diverge?}
\label{sec:quality}
The LLM audit matches the annotators' means on Interface Realism and Outcome
Correctness but is stricter on State Coherence, a state's consistency with its
history.

Six annotators with computer science degrees rated the gold next
states of 200 items for State Coherence, Interface Realism, Outcome
Correctness, and Constraint Enforcement (Appendix~\ref{app:human}). Each annotator rated 100 items, giving 600 ratings in total. We allocated 480 of these to a common set of 80 items, so that every item in that set carries six independent ratings and agreement can be estimated per item, and used the remaining 120 to extend coverage to 200 items across skills and domains. Table~\ref{tab:agreement} gives the agreement scores. Annotators agree on every State Coherence rating and on nearly all Constraint Enforcement and Outcome Correctness ratings, and most of their disagreement is on Interface Realism, where 80.8\% of annotator pairs agree. Most Interface Realism ratings below 5 note that the item uses a probe tool or error format that \datasetname{} injects into the environment.

GPT-5.6 Terra applied the same rubric to 1,627 of
the 1,632 items. On the 78 anchor items in the release, it is stricter than
the annotators on State Coherence (4.73 vs.\ 5.00), where it scores 9 items
below 4 that every annotator gave a 5. In five of the nine, it recomputes a
sum, count, or code from the history and disagrees with the gold, and in the
other four it expects behavior the environment did not show
(Appendix~\ref{app:llm-audit}).

\begin{table}[htbp]
\centering
\caption{Human ratings of the 80 anchor items that all six annotators rated:
mean rating (1--5) and agreement, the share of annotator pairs that give an
item the same score. }
\label{tab:agreement}
\small
\begin{tabular}{@{}lcc@{}}
\toprule
\textbf{Dimension} & \textbf{Mean} & \textbf{Agreement (\%)} \\
\midrule
State Coherence & 5.00 & 100.0 \\
Interface Realism & 4.86 & 80.8 \\
Constraint Enforcement & 5.00 & 97.5 \\
Outcome Correctness & 4.96 & 94.5 \\
\bottomrule
\end{tabular}
\end{table}

\section{Additional analysis}
\label{sec:analysis}

\subsection{Quantitative results}
\label{sec:quant}

\paragraph{Skill profiles.}
\label{sec:skills}
The 19 skills have six distinct leaders (Figure~\ref{fig:tm-heatmap} and Appendix
Figure~\ref{fig:tm-bmw}). GPT-6 Astra leads 11 of them, including 7 of the 9
hardest, and we examine why in Section~\ref{sec:qual}. Of the other eight,
Qwen 3.8 Max leads three, Claude Opus 5.5 two, and Claude Fable 5.1, GPT-5.6
Sol, and Grok 4.6 one each, although Grok 4.6 shares its lead on Rule
Enforcement with two other models (Appendix~\ref{app:skill-results}).
Solvability is not merely asserted by our gates: at least one model recovers the executed answer on 1,001 of the 1,632 items, and on seven skills every item is solved by some model.
All 20 models solve only 4 items (Appendix Table~\ref{tab:solve-hist}). The 631 items that no model solves are mostly L3 (87.6\%) and beyond
40 turns of reach-back (93.7\%), and 85.1\% come from six skills (Appendix
Table~\ref{tab:unsolved}).

\begin{figure}[t]
\centering
\includegraphics[width=0.9\linewidth]{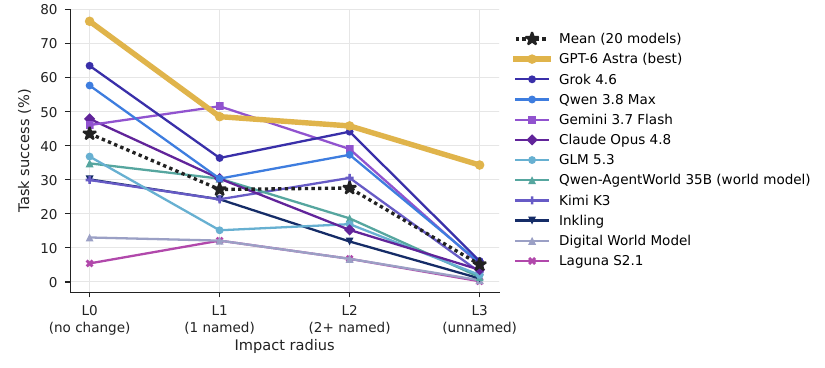}
\caption{Impact radius: task success by level (defined in the text) for 11 of
the 20 models, with the 20-model mean dotted; Appendix Figure~\ref{fig:radius}
shows all 20 models and 95\% intervals for the mean. Every model scores lowest at L3, and GPT-6 Astra's lead over the other models is largest there (Appendix Table~\ref{tab:per-model-state}).}
\label{fig:tm-radius}
\end{figure}

\paragraph{Impact radius and reach-back.}
\label{sec:radius}
A world model has to carry state both across the entities that an action affects and across long stretches of the trajectory. On \datasetname{}, models lose accuracy along both axes, which we measure with impact radius and reach-back.
Accuracy is highest at L0 and lowest at L3 (Figure~\ref{fig:tm-radius}).
Mean success is 43.49\% at L0, where nothing should change. It is nearly the
same at L1 (27.12\%) and L2 (27.54\%), where only entities that the action
names should change, and falls to 5.10\% at L3, where unnamed entities should
also change.
Models may give a plausible value for a named entity but fail to update the
entities that depend on it.
However, almost all L3 items are synthetic long-horizon audits, which are L3 by
construction (Section~\ref{sec:curation}), and we read examples of them in
Section~\ref{sec:qual}.

\begin{figure}[t]
\centering
\includegraphics[width=0.9\linewidth]{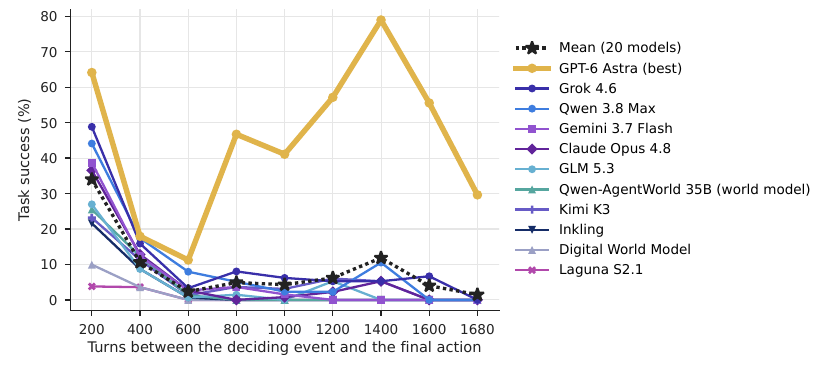}
\caption{Long horizon: task success by reach-back in right-closed 200-turn bins
labeled by their upper edge, for 11 of the 20 models and the 20-model mean.
Items per bin: 764, 195, 151, 137, 129, 133, 19, 45, 27 (Appendix
Table~\ref{tab:bins200}). Beyond 600 turns GPT-6 Astra's curve rises again,
whereas the pooled rate of the other 19 models stays at or below 8.31\% in
every bin beyond 400 turns. Appendix~\ref{app:reachback} relates Astra's rise
to the mix of operators in those bins.}
\label{fig:tm-reachback}
\end{figure}

\label{sec:horizon}%
Every model does worse beyond 40 turns of reach-back than within 40 turns, and
pooled success falls from 42.3\% to 11.4\% (Appendix~\ref{app:difficulty}).
Five skills (Access Control, Schema Conformance, Read and Write Persistence,
Computational Accuracy, and Contradiction Detection) make up 823 of the 1,192
items beyond 40 turns, and their pooled success there ranges from 4.5\% to
11.2\%. In the longer bins of Figure~\ref{fig:tm-reachback}, GPT-6 Astra's curve rises again, whereas the pooled rate of the other 19 models stays at or below 8.31\% in every bin beyond 400 turns. The rise is consistent with the mix of operators in those bins rather than with distance (Appendix~\ref{app:reachback}).

\begin{figure}[t]
\centering
\includegraphics[width=0.92\linewidth]{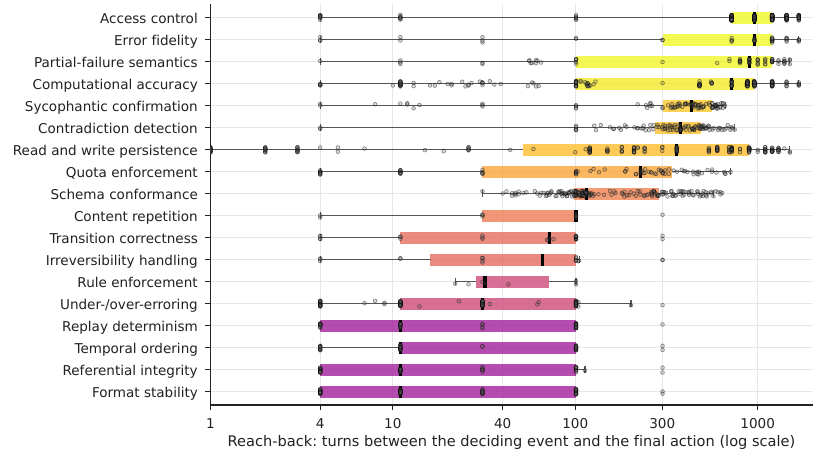}
\caption{Reach-back by primary skill (log scale): quartile boxes, median marked, 1.5$\times$IQR whiskers, jittered items, color by median. Side-Effect Propagation (32 items) has no reach-back and is not shown. Reach-back and skill are confounded by construction: the median reach-back is 11 turns for Format Stability, Temporal Ordering, and Referential Integrity, and 720 to 960 turns for Computational Accuracy, Access Control, and Error Fidelity.}
\label{fig:tm-skill-reachback}
\end{figure}

Reach-back is strongly associated with primary skill
(Figure~\ref{fig:tm-skill-reachback}). The correlation ratio $\eta$, which is 0
when skill says nothing about reach-back and 1 when skill determines it, is
0.69, and Cram\'er's $V$ between skill and reach-back quartile, on the same
scale, is 0.54.

We next test whether the skill mix explains the reach-back drop. If skills alone explained the drop, comparing items with shorter reach-back
(at most 40 turns of reach-back) with longer ones within the same skill would
remove the gap. Instead, the gap shrinks only
from 30.84 to 25.11 points (Appendix Table~\ref{tab:confound}), because in nine
skills the short items are local probes while the long items are audits with
long outputs. The gap disappears when we compare items
from the same operator ($+$1.76 points, 95\% CI $-$1.57 to 5.17) or items
that share both skill and radius ($-$2.20, $-$5.31 to 0.92). However, the comparison within operators rests on the 307
items with longer reach-back that share an operator with shorter ones. Nearly all of them lie within
301 turns, and the audits of 720 to 1,680 turns have no short items to compare
with.
The drop is concentrated in L3 items, where success is about 40\% on the 14
short items but about 4\% on the long ones. L0 items show no drop (42.6\%
short vs.\ 45.7\% long).

GPT-6 Astra passes 34.29\% of L3 items and 40.52\% of items beyond 40 turns, while the next best model on both, Claude Opus 5.5, passes 8.39\% and 16.36\% (Appendix Table~\ref{tab:per-model-state}).

\subsection{Qualitative analysis}
\label{sec:qual}

\paragraph{Where GPT-6 Astra's lead comes from.}
We read 22 items and find that the models differ mostly in the hidden state
they track rather than in how they apply the rules. Most of GPT-6 Astra's lead
over Grok 4.6 comes from long L3 audits that depend on such state
(Appendix~\ref{app:why-astra}).
One of them is the weighted-quorum audit
\texttt{alquorum1-api-\allowbreak s22020000-\allowbreak injected} (Enterprise, L3).
Box~1 in Appendix~\ref{app:quorum} shows it verbatim, and its gold and all 20
models' answers are in the release. Its final action audits requests to
change the lifecycle state of resources. A request is allowed only if its
sponsors' weighted votes reach a threshold and the resource is in the state
that the request expects. The weights and the state are computed from the resource's
capability by formulas stated in the first action, 720 turns before the final
action. The capability is set or changed 49 times in between, but the trajectory
never prints it. On the first request, Astra derives the true
value, 89, reproduces the weights (1/3/1, total 5), and allows the request.
Grok 4.6 applies the same formulas to a capability of 262. Its weights
(4/1/4), its state \texttt{suspended}, and its \texttt{stale\_state} denial all
follow correctly from 262, so the error lies in the tracked state rather than
in the arithmetic. Astra fails 20 of the 76 audits of this kind.

\paragraph{Failure patterns across models.}
\label{sec:anatomy}
Claude Opus 5.5 returns parseable JSON on 80\% of the JSON items it answers,
and GPT-6 Sol on 73\%. Most of their other answers are prose declines
or JSON wrapped in Markdown fences. In contrast, GPT-6 Astra, Grok 4.6,
Qwen 3.8 Max, GPT-5.5, and GPT-5.6 Sol reach 99\% or more, so their failures
are mostly well-formed answers that report a wrong state. In one
game whose earlier turns used up a crafting quota, Qwen 3.8 Max predicted
\texttt{Crafted one charge for banana (2/2)}, whereas the game returned
\texttt{banana is full; crafting quota exceeded}. The prediction is plausible
for the final action alone but ignores a limit set eleven turns earlier
(Appendix~\ref{app:qualitative}).

\paragraph{Specialized text world models.}
Qwen-AgentWorld beats the median of the 18 general-purpose LLMs on 5 of the 19
skills, and DWM on none. One of the five is Side-Effect Propagation, where
Qwen-AgentWorld passes 15 of 32 items. However, the 95\% interval of its margin
over the median includes zero (Appendix~\ref{app:wm-results}), and when we read
those items, we find no side-effect behavior specific to it. Qwen-AgentWorld trails GPT-6 Astra by 35 points. We do not evaluate its base model, so these results do not show how much its training on next-observation prediction changed its accuracy.

\section{Conclusion}
\label{sec:conclusion}

We introduced \datasetname{}, a benchmark of 1,632 items for text world models
in five domains, organized by four capabilities and 19 skills. The best model,
GPT-6 Astra, solves 49.57\% of the items, and 631 items are solved by none of
the 20 models. Accuracy is lowest where an action's effects reach dependent
objects or where reach-back exceeds 40 turns, mostly in long audits that
require tracking hidden state. Many failures are well-formed but report the
wrong state. Because every gold comes from execution and the grader calls a
semantic judge only when deterministic gates cannot decide, \datasetname{}
shows where a simulator departs from its environment. However, each item tests
only one transition after a fixed prefix, so we do not measure how a
simulator's own errors compound over a rollout. Task success is also binary
over the whole next state, so a long audit fails if any single value is wrong.
Finally, since reach-back and impact radius largely follow the skill families that
produced each item (Section~\ref{sec:curation}), they locate hard item families
without showing that distance or breadth causes the errors. We hope
\datasetname{} will lead to more reliable evaluation of text world models and
to better training signals.

\FloatBarrier

\subsection*{AI use statement}
Under the authors' direction, we used generative AI tools, namely LLM coding
agents (Claude Code with Claude models and, earlier, Codex with GPT models). We
used them to implement code for the data pipeline, evaluation harness, and
analyses, and to write the seeded operators that produce the synthetic items,
including their templates and filler turns. They also assisted in analyzing and
interpreting results, including the sample-level analysis in
Section~\ref{sec:qual}, and suggested additional analyses. Every gold next
state was obtained by executing the action against a pinned backend, and none
was produced by an LLM. We did not use AI to propose or refine hypotheses, to
develop the capability and skill taxonomies, or to clean or reformat datasets.
Mathematical claims, proofs, and translation do not apply to this work. We
reviewed all AI-assisted work. We also recomputed every reported number from
the saved outputs with tested scripts, and we checked citations against their
sources. We take responsibility for the final content, including anything
produced with AI.

\subsection*{Ethics statement}
\datasetname{} measures how faithfully language models simulate code,
enterprise, game, math, and dialogue environments. We built its items either from our own
recipes or by adapting interfaces and trajectories from public sources such as
BFCL, TextWorld, $\tau^2$-bench, and SymPy. The release records the source and
license of every item (Appendix~\ref{app:domains}).
In addition to the quality gates, every released item passes an automated
screen for personal data, which rejects real e-mail addresses, personal account
identifiers, and credential-like strings. For the human audit, we recruited six
annotators through Upwork to rate 200 items, and we checked these items for
personally identifying, sensitive, and harmful content
(Appendix~\ref{app:human}).

The main risk we see is misplaced trust in a simulator. If a simulator is wrong
in ways that an agent can exploit, it trains the agent against those mistakes
(Section~\ref{sec:intro}). We built \datasetname{} to expose such errors.
However, since each item tests one transition after a fixed prefix and the
items come from five domains, a high score does not show that a simulator is
reliable for training (Section~\ref{sec:conclusion}).

\subsection*{Reproducibility statement}
We release the benchmark items, evaluation harness, and reproduction
instructions at \url{https://github.com/patronus-ai/globebench}, and the
items and per-model results at
\url{https://huggingface.co/datasets/PatronusAI/globebench}. The GitHub
release also includes the model outputs and judge records, so the results can
be rescored without querying the models again.

Section~\ref{sec:curation} and Appendix~\ref{app:benchmark-details} describe
the pinned backends and how the items were constructed.
Appendix~\ref{app:domains} gives the provenance record of each item, and
Appendix~\ref{app:quality-gates} lists the quality gates.
Section~\ref{sec:setup} and Appendix~\ref{app:evaluation} describe the models
we evaluated and their reasoning, routing, and sampling settings. The same
sections also explain how we handle refusals and how the hybrid grader works,
and they give the judge configuration and prompts.
Appendix~\ref{app:quality-audit} contains the annotation guidelines, the
rubric, and the procedure for the LLM audit. Finally,
Appendix~\ref{app:additional-results} defines the pooled rates, the rounding,
and the bootstrap intervals behind the reported results.

\bibliography{neurips_2026}

\begin{thebibliography}{45}
\providecommand{\natexlab}[1]{#1}
\providecommand{\url}[1]{\texttt{#1}}
\expandafter\ifx\csname urlstyle\endcsname\relax
  \providecommand{\doi}[1]{doi: #1}\else
  \providecommand{\doi}{doi: \begingroup \urlstyle{rm}\Url}\fi

\bibitem[{Alibaba Group}(2026)]{alibaba2026qwen38max}
{Alibaba Group}.
\newblock Alibaba unveils {Qwen3.8-Max}: Its largest and most capable flagship
  model to date, August 2026.
\newblock URL
  \url{https://www.alibabagroup.com/en-US/document-2021044032125272064}.
\newblock Accessed 30 August 2026.

\bibitem[{Anthropic}(2026{\natexlab{a}})]{anthropic2026claudefable5}
{Anthropic}.
\newblock {Claude Fable 5} and {Claude Mythos 5}, June 2026{\natexlab{a}}.
\newblock URL \url{https://www.anthropic.com/news/claude-fable-5-mythos-5}.
\newblock Accessed 23 September 2026.

\bibitem[{Anthropic}(2026{\natexlab{b}})]{anthropic2026claudefable51}
{Anthropic}.
\newblock Introducing {Claude Fable 5.1} and {Claude Mythos 5.1}, September
  2026{\natexlab{b}}.
\newblock URL \url{https://www.anthropic.com/claude-fable-and-mythos-5-1}.
\newblock Accessed 23 September 2026.

\bibitem[{Anthropic}(2026{\natexlab{c}})]{anthropic2026claudeopus48}
{Anthropic}.
\newblock Introducing {Claude Opus 4.8}, May 2026{\natexlab{c}}.
\newblock URL \url{https://www.anthropic.com/news/claude-opus-4-8}.
\newblock Accessed 30 August 2026.

\bibitem[{Anthropic}(2026{\natexlab{d}})]{anthropic2026claudeopus5}
{Anthropic}.
\newblock Introducing {Claude Opus 5}, July 2026{\natexlab{d}}.
\newblock URL \url{https://www.anthropic.com/news/claude-opus-5}.
\newblock Accessed 23 September 2026.

\bibitem[{Anthropic}(2026{\natexlab{e}})]{anthropic2026claudeopus55}
{Anthropic}.
\newblock Introducing {Claude Opus 5.5}, September 2026{\natexlab{e}}.
\newblock URL \url{https://www.anthropic.com/claude-opus-5-5}.
\newblock Accessed 23 September 2026.

\bibitem[{Anthropic}(2026{\natexlab{f}})]{anthropic2026cyber}
{Anthropic}.
\newblock More details on {Fable 5}'s cyber safeguards and our jailbreak
  framework, July 2026{\natexlab{f}}.
\newblock URL
  \url{https://www.anthropic.com/news/fable-safeguards-jailbreak-framework}.
\newblock Accessed 25 September 2026.

\bibitem[{Anthropic}(2026{\natexlab{g}})]{anthropic2026refusals}
{Anthropic}.
\newblock Refusals and fallback, 2026{\natexlab{g}}.
\newblock URL
  \url{https://platform.claude.com/docs/en/build-with-claude/refusals-and-fallback}.
\newblock Claude Platform Docs. Accessed 25 September 2026.

\bibitem[{Astropy Collaboration} et~al.(2022){Astropy Collaboration},
  Price-Whelan, Lim, Earl, Starkman, Bradley, Shupe, Patil, Corrales, Brasseur,
  et~al.]{astropy2022astropy}
{Astropy Collaboration}, Adrian~M. Price-Whelan, Pey~Lian Lim, Nicholas Earl,
  Nathaniel Starkman, Larry Bradley, David~L. Shupe, Aarya~A. Patil, Lia
  Corrales, C.~E. Brasseur, et~al.
\newblock The {Astropy} project: Sustaining and growing a community-oriented
  open-source project and the latest major release (v5.0) of the core package.
\newblock \emph{The Astrophysical Journal}, 935\penalty0 (2):\penalty0 167,
  2022.
\newblock \doi{10.3847/1538-4357/ac7c74}.

\bibitem[Barres et~al.(2025)Barres, Dong, Ray, Si, and
  Narasimhan]{barres2025tau2}
Victor Barres, Honghua Dong, Soham Ray, Xujie Si, and Karthik Narasimhan.
\newblock {$\tau^2$-Bench}: Evaluating conversational agents in a dual-control
  environment.
\newblock \emph{arXiv preprint arXiv:2506.07982}, 2025.
\newblock URL \url{https://arxiv.org/abs/2506.07982}.

\bibitem[Cock et~al.(2009)Cock, Antao, Chang, Chapman, Cox, Dalke, Friedberg,
  Hamelryck, Kauff, Wilczynski, and de~Hoon]{cock2009biopython}
Peter J.~A. Cock, Tiago Antao, Jeffrey~T. Chang, Brad~A. Chapman, Cymon~J. Cox,
  Andrew Dalke, Iddo Friedberg, Thomas Hamelryck, Frank Kauff, Bartek
  Wilczynski, and Michiel J.~L. de~Hoon.
\newblock {Biopython}: Freely available {Python} tools for computational
  molecular biology and bioinformatics.
\newblock \emph{Bioinformatics}, 25\penalty0 (11):\penalty0 1422--1423, 2009.
\newblock \doi{10.1093/bioinformatics/btp163}.

\bibitem[C{\^o}t{\'e} et~al.(2018)C{\^o}t{\'e}, K{\'a}d{\'a}r, Yuan, Kybartas,
  Barnes, Fine, Moore, Tao, Hausknecht, {El Asri}, Adada, Tay, and
  Trischler]{cote18textworld}
Marc-Alexandre C{\^o}t{\'e}, {\'A}kos K{\'a}d{\'a}r, Xingdi Yuan, Ben Kybartas,
  Tavian Barnes, Emery Fine, James Moore, Ruo~Yu Tao, Matthew Hausknecht, Layla
  {El Asri}, Mahmoud Adada, Wendy Tay, and Adam Trischler.
\newblock {TextWorld}: A learning environment for text-based games.
\newblock \emph{arXiv preprint arXiv:1806.11532}, 2018.
\newblock URL \url{https://arxiv.org/abs/1806.11532}.

\bibitem[Deshpande(2026)]{deshpande2026masked}
Darshan Deshpande.
\newblock Masked diffusion language models are strong and steerable text-based
  world models for agentic {RL}.
\newblock \emph{arXiv preprint arXiv:2607.16204}, 2026.

\bibitem[Fiekas(2025)]{fiekas2025pythonchess}
Niklas Fiekas.
\newblock python-chess: A chess library for {Python}, 2025.
\newblock URL \url{https://github.com/niklasf/python-chess}.
\newblock Version 1.11.2. Accessed September 2026.

\bibitem[Goodwin et~al.(2024)Goodwin, Moffat, Schoegl, Speth, and
  Weber]{goodwin2024cantera}
David~G. Goodwin, Harry~K. Moffat, Ingmar Schoegl, Raymond~L. Speth, and
  Bryan~W. Weber.
\newblock Cantera: An object-oriented software toolkit for chemical kinetics,
  thermodynamics, and transport processes, 2024.
\newblock URL \url{https://www.cantera.org}.
\newblock Version 3.1.0.

\bibitem[{Google DeepMind}(2026)]{googledeepmind2026gemini37flash}
{Google DeepMind}.
\newblock {Gemini 3.7 Flash} model card, August 2026.
\newblock URL
  \url{https://deepmind.google/models/model-cards/gemini-3-7-flash/}.
\newblock Accessed 30 August 2026.

\bibitem[Hu et~al.(2025)Hu, Xiong, Yi, Wei, Xiao, Chen, Ye, Tao, Zhou, Zhao,
  Li, Xu, Wang, Xu, Qiao, Wang, Kuang, Zeng, Wang, Li, Jiang, Zhou, Wang, Yin,
  Zhao, Yang, Wu, Zhang, and Wu]{hu-etal-2025-os}
Xueyu Hu, Tao Xiong, Biao Yi, Zishu Wei, Ruixuan Xiao, Yurun Chen, Jiasheng Ye,
  Meiling Tao, Xiangxin Zhou, Ziyu Zhao, Yuhuai Li, Shengze Xu, Shenzhi Wang,
  Xinchen Xu, Shuofei Qiao, Zhaokai Wang, Kun Kuang, Tieyong Zeng, Liang Wang,
  Jiwei Li, Yuchen~Eleanor Jiang, Wangchunshu Zhou, Guoyin Wang, Keting Yin,
  Zhou Zhao, Hongxia Yang, Fan Wu, Shengyu Zhang, and Fei Wu.
\newblock {OS} agents: A survey on {MLLM}-based agents for computer, phone and
  browser use.
\newblock In Wanxiang Che, Joyce Nabende, Ekaterina Shutova, and Mohammad~Taher
  Pilehvar (eds.), \emph{Proceedings of the 63rd Annual Meeting of the
  Association for Computational Linguistics (Volume 1: Long Papers)}, pp.\
  7436--7465, Vienna, Austria, July 2025. Association for Computational
  Linguistics.
\newblock ISBN 979-8-89176-251-0.
\newblock \doi{10.18653/v1/2025.acl-long.369}.
\newblock URL \url{https://aclanthology.org/2025.acl-long.369/}.

\bibitem[Jain et~al.(2025)Jain, Singh, Shetty, Zheng, Sen, and
  Stoica]{jain2025r2e}
Naman Jain, Jaskirat Singh, Manish Shetty, Liang Zheng, Koushik Sen, and Ion
  Stoica.
\newblock {R2E-Gym}: Procedural environments and hybrid verifiers for scaling
  open-weights {SWE} agents.
\newblock \emph{arXiv preprint arXiv:2504.07164}, 2025.

\bibitem[{Jakumetsu}(2025)]{jakumetsu2025mcpmarklogs}
{Jakumetsu}.
\newblock {MCPMark} trajectory logs (mcpmark-v1-0905).
\newblock Hugging Face dataset, 2025.
\newblock URL
  \url{https://huggingface.co/datasets/Jakumetsu/mcpmark-trajectory-log}.
\newblock MIT license. Accessed September 2026.

\bibitem[Jimenez et~al.(2024)Jimenez, Yang, Wettig, Yao, Pei, Press, and
  Narasimhan]{jimenez2023swe}
Carlos~E. Jimenez, John Yang, Alexander Wettig, Shunyu Yao, Kexin Pei, Ofir
  Press, and Karthik Narasimhan.
\newblock {SWE-bench}: Can language models resolve real-world {GitHub} issues?
\newblock In \emph{International Conference on Learning Representations}, 2024.
\newblock URL
  \url{https://proceedings.iclr.cc/paper_files/paper/2024/hash/edac78c3e300629acfe6cbe9ca88fb84-Abstract-Conference.html}.

\bibitem[Lanctot et~al.(2019)Lanctot, Lockhart, Lespiau, Zambaldi, Upadhyay,
  P{\'e}rolat, Srinivasan, Timbers, Tuyls, Omidshafiei, Hennes, Morrill,
  Muller, Ewalds, Faulkner, Kram{\'a}r, De~Vylder, Saeta, Bradbury, Ding,
  Borgeaud, Lai, Schrittwieser, Anthony, Hughes, Danihelka, and
  Ryan-Davis]{lanctot2019openspiel}
Marc Lanctot, Edward Lockhart, Jean-Baptiste Lespiau, Vinicius Zambaldi,
  Satyaki Upadhyay, Julien P{\'e}rolat, Sriram Srinivasan, Finbarr Timbers,
  Karl Tuyls, Shayegan Omidshafiei, Daniel Hennes, Dustin Morrill, Paul Muller,
  Timo Ewalds, Ryan Faulkner, J{\'a}nos Kram{\'a}r, Bart De~Vylder, Brennan
  Saeta, James Bradbury, David Ding, Sebastian Borgeaud, Matthew Lai, Julian
  Schrittwieser, Thomas Anthony, Edward Hughes, Ivo Danihelka, and Jonah
  Ryan-Davis.
\newblock {OpenSpiel}: A framework for reinforcement learning in games.
\newblock \emph{arXiv preprint arXiv:1908.09453}, 2019.
\newblock URL \url{https://arxiv.org/abs/1908.09453}.

\bibitem[Landrum et~al.(2026)]{landrum2026rdkit}
Greg Landrum et~al.
\newblock {RDKit}: Open-source cheminformatics, 2026.
\newblock URL \url{https://www.rdkit.org}.
\newblock Version 2026.03.5. Accessed September 2026.

\bibitem[Li et~al.(2026)Li, Wang, Qiu, Yin, Zhang, Qian, Li, Ma, Chen, and
  Ji]{li2025word}
Yixia Li, Hongru Wang, Jiahao Qiu, Zhenfei Yin, Dongdong Zhang, Cheng Qian,
  Zeping Li, Xiaoteng Ma, Guanhua Chen, and Heng Ji.
\newblock From word to world: Can large language models be implicit text-based
  world models?
\newblock In \emph{Proceedings of the 64th Annual Meeting of the Association
  for Computational Linguistics (Volume 1: Long Papers)}, pp.\  8084--8111, San
  Diego, California, United States, 2026. Association for Computational
  Linguistics.
\newblock \doi{10.18653/v1/2026.acl-long.366}.
\newblock URL \url{https://aclanthology.org/2026.acl-long.366/}.

\bibitem[Liu et~al.(2026)Liu, Hui, Zhang, Sun, Su, Wang, and
  Su]{liu2026envsimbenchbenchmarkevaluatingimproving}
Yi~Liu, TingFeng Hui, Wei Zhang, Li~Sun, Ningxin Su, Jian Wang, and Sen Su.
\newblock {EnvSimBench}: A benchmark for evaluating and improving {LLM}-based
  environment simulation.
\newblock \emph{arXiv preprint arXiv:2605.07247}, 2026.
\newblock URL \url{https://arxiv.org/abs/2605.07247}.

\bibitem[L{\"u}nsdorf \& Scherfke(2026)L{\"u}nsdorf and
  Scherfke]{lunsdorf2026simpy}
Ontje L{\"u}nsdorf and Stefan Scherfke.
\newblock {SimPy}: Discrete event simulation for {Python}, 2026.
\newblock URL \url{https://gitlab.com/team-simpy/simpy}.
\newblock Version 4.1.2. Accessed September 2026.

\bibitem[Meurer et~al.(2017)Meurer, Smith, Paprocki, \v{C}ert\'{i}k, Kirpichev,
  Rocklin, Kumar, Ivanov, Moore, Singh, Rathnayake, Vig, Granger, Muller,
  Bonazzi, Gupta, Vats, Johansson, Pedregosa, Curry, Terrel, Rou\v{c}ka, Saboo,
  Fernando, Kulal, Cimrman, and Scopatz]{10.7717/peerj-cs.103}
Aaron Meurer, Christopher~P. Smith, Mateusz Paprocki, Ond\v{r}ej
  \v{C}ert\'{i}k, Sergey~B. Kirpichev, Matthew Rocklin, AMiT Kumar, Sergiu
  Ivanov, Jason~K. Moore, Sartaj Singh, Thilina Rathnayake, Sean Vig, Brian~E.
  Granger, Richard~P. Muller, Francesco Bonazzi, Harsh Gupta, Shivam Vats,
  Fredrik Johansson, Fabian Pedregosa, Matthew~J. Curry, Andy~R. Terrel,
  \v{S}t\v{e}p\'{a}n Rou\v{c}ka, Ashutosh Saboo, Isuru Fernando, Sumith Kulal,
  Robert Cimrman, and Anthony Scopatz.
\newblock {SymPy}: symbolic computing in {Python}.
\newblock \emph{PeerJ Computer Science}, 3:\penalty0 e103, January 2017.
\newblock ISSN 2376-5992.
\newblock \doi{10.7717/peerj-cs.103}.
\newblock URL \url{https://doi.org/10.7717/peerj-cs.103}.

\bibitem[{Moonshot AI}(2026)]{moonshotai2026kimik3}
{Moonshot AI}.
\newblock {Kimi K3}: Open frontier intelligence, July 2026.
\newblock URL \url{https://www.kimi.com/en/blog/kimi-k3}.
\newblock Accessed 30 August 2026.

\bibitem[{OpenAI}(2026{\natexlab{a}})]{openai2026gpt55}
{OpenAI}.
\newblock Introducing {GPT-5.5}, April 2026{\natexlab{a}}.
\newblock URL \url{https://openai.com/index/introducing-gpt-5-5/}.
\newblock Accessed 30 August 2026.

\bibitem[{OpenAI}(2026{\natexlab{b}})]{openai2026gpt56}
{OpenAI}.
\newblock {GPT-5.6}: Frontier intelligence that scales with your ambition, July
  2026{\natexlab{b}}.
\newblock URL \url{https://openai.com/index/gpt-5-6/}.
\newblock Accessed 30 August 2026.

\bibitem[{OpenAI}(2026{\natexlab{c}})]{openai2026gpt6astra}
{OpenAI}.
\newblock {GPT-6 Astra}: A new generation of intelligence, September
  2026{\natexlab{c}}.
\newblock URL \url{https://openai.com/index/gpt-6-astra/}.
\newblock Accessed 23 September 2026.

\bibitem[{OpenAI}(2026{\natexlab{d}})]{openai2026gpt6solluna}
{OpenAI}.
\newblock Introducing {GPT-6 Sol} and {Luna}, September 2026{\natexlab{d}}.
\newblock URL \url{https://openai.com/index/introducing-gpt-6-sol-and-luna/}.
\newblock Accessed 23 September 2026.

\bibitem[Patil et~al.(2025)Patil, Mao, Yan, Ji, Suresh, Stoica, and
  Gonzalez]{patil2025bfcl}
Shishir~G. Patil, Huanzhi Mao, Fanjia Yan, Charlie Cheng-Jie Ji, Vishnu Suresh,
  Ion Stoica, and Joseph~E. Gonzalez.
\newblock The {Berkeley Function Calling Leaderboard (BFCL)}: From tool use to
  agentic evaluation of large language models.
\newblock In \emph{Proceedings of the 42nd International Conference on Machine
  Learning}, volume 267 of \emph{Proceedings of Machine Learning Research},
  pp.\  48371--48392, 2025.
\newblock URL \url{https://proceedings.mlr.press/v267/patil25a.html}.

\bibitem[{Patronus AI}(2026)]{patronus2026dwm}
{Patronus AI}.
\newblock Announcing our \$50{M} series {B} to simulate the entire world's
  intelligence and unveiling our first {Digital World Model} for {AI} agent
  training and simulation, June 2026.
\newblock URL
  \url{https://www.patronus.ai/announcements/announcing-our-50m-series-b}.
\newblock Accessed 24 September 2026.

\bibitem[{Poolside}(2026)]{poolside2026lagunas21}
{Poolside}.
\newblock Introducing {Laguna S 2.1}, July 2026.
\newblock URL \url{https://poolside.ai/blog/introducing-laguna-s-2-1}.
\newblock Accessed 30 August 2026.

\bibitem[Qiang et~al.(2025)Qiang, Zhuang, Li, Sagar V~K, Zhang, Li, Wong, Yang,
  Liang, Zhang, and Dai]{qiang2026mle}
Rushi Qiang, Yuchen Zhuang, Yinghao Li, Dingu Sagar V~K, Rongzhi Zhang,
  ChangHao Li, Ian Wong, Sherry Yang, Percy Liang, Chao Zhang, and Bo~Dai.
\newblock {MLE-Dojo}: Interactive environments for empowering {LLM} agents in
  machine learning engineering.
\newblock In \emph{Advances in Neural Information Processing Systems},
  volume~38, 2025.
\newblock \doi{10.52202/085713-0139}.
\newblock URL
  \url{https://proceedings.neurips.cc/paper_files/paper/2025/hash/0603c69125ad4b964bc9c4832f7b9f8f-Abstract-Datasets_and_Benchmarks_Track.html}.

\bibitem[{Qwen Team}(2026)]{qwen2026qwen38flashnext}
{Qwen Team}.
\newblock {Qwen3.8-Flash-Next}: A new architecture, towards ultimate
  cost-efficiency, August 2026.
\newblock URL \url{https://qwen.ai/blog?id=qwen3.8-flash-next}.
\newblock Accessed 23 September 2026.

\bibitem[{SpaceXAI}(2026)]{spacexai2026grok46}
{SpaceXAI}.
\newblock Introducing {Grok 4.6}, August 2026.
\newblock URL \url{https://x.ai/news/grok-4-6}.
\newblock Accessed 30 August 2026.

\bibitem[{Thinking Machines Lab}(2026)]{thinkingmachines2026inkling}
{Thinking Machines Lab}.
\newblock Inkling: Our open-weights model, July 2026.
\newblock URL \url{https://thinkingmachines.ai/news/introducing-inkling/}.
\newblock Accessed 30 August 2026.

\bibitem[Virtanen et~al.(2020)Virtanen, Gommers, Oliphant, Haberland, Reddy,
  Cournapeau, Burovski, Peterson, Weckesser, Bright, {van der Walt}, Brett,
  Wilson, Millman, Mayorov, Nelson, Jones, Kern, Larson, Carey, Polat, Feng,
  Moore, {VanderPlas}, Laxalde, Perktold, Cimrman, Henriksen, Quintero, Harris,
  Archibald, Ribeiro, Pedregosa, {van Mulbregt}, and {SciPy 1.0
  Contributors}]{virtanen2020scipy}
Pauli Virtanen, Ralf Gommers, Travis~E. Oliphant, Matt Haberland, Tyler Reddy,
  David Cournapeau, Evgeni Burovski, Pearu Peterson, Warren Weckesser, Jonathan
  Bright, St{\'e}fan~J. {van der Walt}, Matthew Brett, Joshua Wilson, K.~Jarrod
  Millman, Nikolay Mayorov, Andrew R.~J. Nelson, Eric Jones, Robert Kern, Eric
  Larson, C~J Carey, {\.I}lhan Polat, Yu~Feng, Eric~W. Moore, Jake
  {VanderPlas}, Denis Laxalde, Josef Perktold, Robert Cimrman, Ian Henriksen,
  E.~A. Quintero, Charles~R. Harris, Anne~M. Archibald, Ant{\^o}nio~H. Ribeiro,
  Fabian Pedregosa, Paul {van Mulbregt}, and {SciPy 1.0 Contributors}.
\newblock {SciPy} 1.0: Fundamental algorithms for scientific computing in
  {Python}.
\newblock \emph{Nature Methods}, 17\penalty0 (3):\penalty0 261--272, 2020.
\newblock \doi{10.1038/s41592-019-0686-2}.

\bibitem[Wu et~al.(2026)Wu, Liu, Zhang, Chen, Meng, Du, Zhao, Zhang, Ye, Wang,
  Wang, Ni, Yang, Xu, and Shieh]{wu2026mcpmark}
Zijian Wu, Xiangyan Liu, Xinyuan Zhang, Lingjun Chen, Fanqing Meng, Lingxiao
  Du, Yiran Zhao, Fanshi Zhang, Yaoqi Ye, Jiawei Wang, Zirui Wang, Jinjie Ni,
  Yufan Yang, Arvin Xu, and Michael~Qizhe Shieh.
\newblock {MCPMark}: A benchmark for stress-testing realistic and comprehensive
  {MCP} use.
\newblock In \emph{The Fourteenth International Conference on Learning
  Representations}, 2026.
\newblock URL \url{https://openreview.net/forum?id=uobROwBsJm}.

\bibitem[Xie et~al.(2024)Xie, Zhang, Chen, Li, Zhao, Cao, Hua, Cheng, Shin,
  Lei, Liu, Xu, Zhou, Savarese, Xiong, Zhong, and Yu]{OSWorld}
Tianbao Xie, Danyang Zhang, Jixuan Chen, Xiaochuan Li, Siheng Zhao, Ruisheng
  Cao, Toh~Jing Hua, Zhoujun Cheng, Dongchan Shin, Fangyu Lei, Yitao Liu,
  Yiheng Xu, Shuyan Zhou, Silvio Savarese, Caiming Xiong, Victor Zhong, and Tao
  Yu.
\newblock {OSWorld}: Benchmarking multimodal agents for open-ended tasks in
  real computer environments.
\newblock In \emph{Advances in Neural Information Processing Systems},
  volume~37, pp.\  52040--52094, 2024.
\newblock \doi{10.52202/079017-1650}.
\newblock URL
  \url{https://proceedings.neurips.cc/paper_files/paper/2024/hash/5d413e48f84dc61244b6be550f1cd8f5-Abstract-Datasets_and_Benchmarks_Track.html}.

\bibitem[Xu et~al.(2026)Xu, Zhang, Yu, Zhang, Meng, Chu, and Tu]{xu2026scaling}
Yucheng Xu, Keyi Zhang, Yuyang Yu, Min Zhang, Shiyuan Meng, Pei Chu, and
  Zhongying Tu.
\newblock Scaling scientific discovery environments for turn-level agentic
  {RL}.
\newblock \emph{arXiv preprint arXiv:2607.28990}, 2026.

\bibitem[Yang et~al.(2025)Yang, Lieret, Jimenez, Wettig, Khandpur, Zhang, Hui,
  Press, Schmidt, and Yang]{yang2025swesmith}
John Yang, Kilian Lieret, Carlos~E. Jimenez, Alexander Wettig, Kabir Khandpur,
  Yanzhe Zhang, Binyuan Hui, Ofir Press, Ludwig Schmidt, and Diyi Yang.
\newblock {SWE-smith}: Scaling data for software engineering agents.
\newblock In \emph{Advances in Neural Information Processing Systems},
  volume~38, 2025.
\newblock \doi{10.52202/085713-3239}.
\newblock URL
  \url{https://proceedings.neurips.cc/paper_files/paper/2025/hash/8b86cf5ace600c48fd188efbb8dedec8-Abstract-Datasets_and_Benchmarks_Track.html}.

\bibitem[{Z.ai}(2026)]{zai2026glm53}
{Z.ai}.
\newblock {GLM-5.3}: Frontier coding with emergent cyber capabilities, August
  2026.
\newblock URL \url{https://z.ai/blog/glm-5.3}.
\newblock Accessed 30 August 2026.

\bibitem[Zuo et~al.(2026)Zuo, Xiao, Sheng, Huang, Tu, Liu, Tang, Hu, Su, Lan,
  et~al.]{zuo2026qwen}
Yuxin Zuo, Zikai Xiao, Li~Sheng, Fei Huang, Jianhong Tu, Yuxuan Liu, Tianyi
  Tang, Xiaomeng Hu, Yang Su, Qingfeng Lan, et~al.
\newblock {Qwen-AgentWorld}: Language world models for general agents.
\newblock \emph{arXiv preprint arXiv:2606.24597}, 2026.
\newblock URL \url{https://arxiv.org/abs/2606.24597}.

\end{thebibliography}
\bibliographystyle{iclr2027_conference}

\clearpage
\appendix

\section{Benchmark construction}
\label{app:benchmark-details}

\subsection{Capabilities and skills}
\label{app:skills}

Table~\ref{tab:capabilities} defines the four capabilities, and
Table~\ref{tab:skills-app} lists the 19 skills. We use the capabilities to
describe at a high level what a realistic and faithful environment must do,
and the skills to make each capability testable. For each skill, the table
gives its construction axes, which are the properties we vary when we build
its items. It also gives the number of items that test the skill and the
capabilities it maps to.

Each item is built to test one primary skill, and it may also carry capability
and failure labels. Because one item can carry several labels, the slices
defined by these labels overlap, and their counts should not be added together.
Figure~\ref{fig:library-skills} in the Introduction gives one example of each
skill. Each example pairs a user turn with the environment's correct
response (Under- and Over-Erroring shows two such pairs), and all of the examples are set in the same library database.

\begin{table}[htbp]
\centering
\caption{The four capabilities. Let $s_t$ be the predicted world state at step
$t$, $a_t$ the agent action, and $h_t=(s_0,a_0,\ldots,a_{t-1})$ the interaction
history.}
\label{tab:capabilities}
\small
\begin{tabular}{@{}>{\raggedright\arraybackslash}p{0.25\linewidth}p{0.7\linewidth}@{}}
\toprule
\textbf{Capability} & \textbf{Definition} \\
\midrule
State Coherence (SC) & Predicted states stay consistent with the entities,
objects, and facts from earlier interactions with the user and the
environment, and every entity they refer to exists in the visible context. \\
\addlinespace
Causal Correctness (CC) & Different combinations of action $a_t$, history $h_t$,
and entities lead to different next states $s_{t+1}$, and each of these states
is correct. A model that maps several combinations to the same state has not
captured how the environment changes. The side effects of an action appear in
later states. \\
\addlinespace
Interface Fidelity (IF) & The predicted next state gives the same level of
detail that the real system would return, with nothing extra and nothing
missing, and it uses the real schema and error format. \\
\addlinespace
Constraint Enforcement (CE) & Sandbox limits on tool access, actions, and states
are respected. Valid actions execute, and invalid or out-of-scope actions raise
the correct error. \\
\bottomrule
\end{tabular}
\end{table}

\begingroup
\setlength{\tabcolsep}{2.5pt}
\renewcommand{\arraystretch}{0.76}
\begin{longtable}{@{}>{\scriptsize\raggedright\arraybackslash}p{0.17\linewidth}>{\scriptsize\raggedright\arraybackslash}p{0.28\linewidth}>{\scriptsize\raggedright\arraybackslash}p{0.25\linewidth}>{\scriptsize}r>{\scriptsize\raggedright\arraybackslash}p{0.09\linewidth}@{}}
\caption{The 19 primary skills, one per row. Description: the behavior that
a faithful environment shows. Axes: the construction axes, the properties we
vary when we build the skill's items. $N$: the number of items whose primary
skill it is (the column sums to 1,632). Cap.: the capabilities that the skill
maps to (Section~\ref{sec:curation}). SC: State
Coherence; CC: Causal Correctness; IF: Interface Fidelity; CE: Constraint
Enforcement.}
\label{tab:skills-app}\\
\toprule
\textbf{Skill} & \textbf{Description} & \textbf{Axes} & \textbf{$N$} & \textbf{Cap.} \\
\midrule
\endfirsthead
\multicolumn{5}{@{}l}{\small\itshape Table \thetable\ continued}\\
\toprule
\textbf{Skill} & \textbf{Description} & \textbf{Axes} & \textbf{$N$} & \textbf{Cap.} \\
\midrule
\endhead
\midrule
\multicolumn{5}{r@{}}{\small\itshape continued on next page}\\
\endfoot
\bottomrule
\endlastfoot

Read and Write Persistence &
Written values remain unchanged until explicitly overwritten. &
Gap length, values tracked, mutation, location & 207 & SC, CC \\

Referential Integrity &
Identifiers, keys, and numerical references resolve to the correct object. &
Number of children, nesting depth & 50 & SC, CC, IF \\

Contradiction Detection &
The same object returns a consistent value through different access paths. &
Path pair, question distance, interleaved change & 110 & SC \\

Replay Determinism (seeded) &
The predicted state matches the canonical state produced by replaying a fixed seed and action sequence. &
Trajectory length, replay point, deterministic control & 41 & SC, CC \\

Sycophantic Confirmation &
False assertions are corrected or refused rather than accepted. &
Repetition count, truth distance & 92 & SC, CE \\

Transition Correctness &
The predicted state reflects the actual effects of the action. &
State-change magnitude, no-op presence & 25 & CC \\

Temporal Ordering and Consistency &
Time advances without frozen clocks, drift, or ignored expiry. &
Injection depth, timestamp format, failure trigger & 43 & SC, CC, IF \\

Computational Accuracy &
Deterministic arithmetic, counts, sorts, edits, and encodings are exact. &
Injection depth, operation type, size complexity & 197 & SC, CC, IF \\

Side-Effect Propagation &
A mutation updates every affected secondary surface. &
Side surface, later steps, explicit prompting & 32 & SC, CC, CE \\

Irreversibility Handling &
Destructive effects persist unless a valid inverse action restores state. &
Operation pair, separation, size, reversibility & 18 & SC, CC \\

Partial-Failure Semantics &
Atomic operations roll back on failure; non-atomic operations retain completed effects. &
Batch size, failure position, atomicity, cause & 71 & CC, CE, IF \\

Schema Conformance &
Responses use the real fields, types, enumerations, and required keys. &
Schema size, nesting, enums, optional fields & 200 & CE, IF \\

Error Fidelity &
A failed action returns the correct error class and referent. &
Error family, confusable pair & 41 & CE, IF \\

Content Repetition &
Distinct values established in the trajectory remain distinct rather than being copied or collapsed. &
Field, sample count, available context & 11 & SC, IF \\

Format Stability &
Response structure remains stable across turns, wrappers, and rephrasing. &
Wrapper type, trajectory position, output length & 69 & SC, IF \\

Rule Enforcement &
A precondition established earlier is enforced by the later transition. &
Injection depth, violation, precondition, domain & 7 & SC, CE \\

Access Control &
Actions outside the active permission, ownership, or tenant boundary are refused. &
Permission boundary, ownership, tenant scope & 224 & SC, CE, IF \\

Quota Enforcement &
Usage is accumulated across the trajectory and checked against the active limit. &
Quota type, probe proximity & 82 & SC, CC, CE \\

Under- and Over-Erroring &
The environment fails exactly when the real system would and accepts valid actions. &
Impossibility cause, gap length, valid twin & 112 & CC, CE \\

\end{longtable}
\endgroup

\subsection{Domains and source data}
\label{app:domains}

The five domains are different kinds of environment that deployed agents act
in. An item may carry several labels, but it belongs to exactly one domain. The
domain determines which backends can execute the item and which actions the
agent can take. It also fixes how the state is represented and which errors the system
raises. Table~\ref{tab:domains} lists the state surface of each domain, the
backends that execute its items, and its number of items. Code is the largest
domain (526 items, 32.2\%) and Dialogue the smallest (193 items, 11.8\%).

To obtain the gold next state of an item, we execute its final action against a
pinned backend, whose software versions are fixed. A backend that runs in a
container is fixed by the digest of its image and runs without network access.
A library backend that runs inside our harness is fixed instead by the hashes
of its dependencies or lock files. Of the 1,137 items generated from our own recipes
(Table~\ref{tab:sources}), the 955 from our synthetic generation campaigns run
on simulators inside our harness. The simulators implement the
rules we wrote for these items.

\begin{table}[htbp]
\caption{Domains, state surfaces, backends supported by the operators, and
released item counts. The state surface is the part of the environment
state that the domain's actions read and change. Items: number of released
items in the domain. Share: Items as a percentage of all 1,632 items.}
\label{tab:domains}
\centering
\scriptsize
\setlength{\tabcolsep}{2.5pt}
\renewcommand{\arraystretch}{1.05}
\begin{tabular}{@{}>{\raggedright\arraybackslash}p{0.11\linewidth}>{\raggedright\arraybackslash}p{0.3\linewidth}>{\raggedright\arraybackslash}p{0.4\linewidth}rr@{}}
\toprule
\textbf{Domain} & \textbf{State surface} & \textbf{Backends} & \textbf{Items} & \textbf{Share} \\
\midrule
Code &
Software repositories, POSIX shells over a filesystem, and databases: file
versions, records, and the commands, edits, and queries that mutate them. &
Alpine 3.19 container; SQLite, TinyDB, fakeredis, and K\`uzu; Git repositories
exercised with pytest inside a pinned image &
526 & 32.2\% \\
\addlinespace
Enterprise &
Business systems driven through workflows and typed API calls: records, roles,
tables, views, and quota counters behind a schema. &
Odoo 17 Community over PostgreSQL 15 via XML-RPC, Beancount ledgers, Tryton,
DuckDB for analytics, the \texttt{transitions} state-machine library,
stateful BFCL v3/v4 tool-use classes, and SimPy schedules &
381 & 23.3\% \\
\addlinespace
Games &
Rule-based text worlds: inventory, location, board state, and turn-by-turn
world dynamics. &
TextWorld (Z-machine), python-chess, Gymnasium toy-text environments, and
OpenSpiel (tic-tac-toe, Connect Four, Breakthrough) &
334 & 20.5\% \\
\addlinespace
Math and Science &
Symbolic and numeric computation in a persistent namespace; every action has an
exact result. &
SymPy; SciPy ODE integrators for physics (pendulum, RC circuit, planar
orbit), chemistry (Brusselator, Michaelis--Menten), and population and neural
dynamics (Lotka--Volterra, SIR, SEIR, FitzHugh--Nagumo); SimPy schedules over
measurements from SciPy, Cantera, RDKit, Astropy, and Biopython &
198 & 12.1\% \\
\addlinespace
Dialogue &
Task-oriented multi-turn conversation: definitions, requests, and commitments
made so far; the world must hold the user to them rather than agree with
whatever is asserted. &
$\tau^2$-bench domain environments backed by a typed state tree loaded from a
JSON seed &
193 & 11.8\% \\
\midrule
\multicolumn{3}{r}{\textbf{Total}} & \textbf{1,632} & \textbf{100.0\%} \\
\bottomrule
\end{tabular}
\end{table}

\paragraph{Sources and licenses.}
Of the 1,632 items, 1,137 are generated from our own recipes and record no
upstream source. The other 495 either adapt data or structure from an upstream
project or take their observations from an upstream library
(Table~\ref{tab:sources}). Most of them come from BFCL (203 items) and
TextWorld (194 items). Every released item stores its source and license
in two metadata fields, \texttt{source} and \texttt{source\_license}.

What counts as the source depends on the backend. For the BFCL, TextWorld,
$\tau^2$-bench, and MCPMark backends, it is the project whose data or
structure the item adapts. For the SymPy, SimPy, OpenSpiel, and python-chess
backends, it is instead the library that computes each observation. One SymPy
item also replays a SymPy doctest. The remaining items do not record their
execution backend, such as Odoo or the Alpine shell, as a source.

The 11 laboratory-schedule items combine SimPy with measurements computed by
SciPy~\citep{virtanen2020scipy}, Cantera~\citep{goodwin2024cantera},
RDKit~\citep{landrum2026rdkit}, Astropy~\citep{astropy2022astropy}, and
Biopython~\citep{cock2009biopython}. Each of these items carries the license
label \texttt{mixed-permissive-see-contract}. The label covers SimPy under MIT,
Biopython under the Biopython License Agreement, and SciPy, Cantera, RDKit,
and Astropy under BSD-3-Clause.

A single item records the GPL-3.0-or-later license. It replays a seeded chess
game of 112 plies in python-chess. The library computes every position,
legal-move count, and check flag in the game. The item's tool description also
quotes the format that python-chess uses for illegal-move errors. Each of the
495 items records the license of its upstream source, and we release our
harness code under Apache-2.0.

\begin{table}[htbp]
\caption{Upstream sources recorded by 495 of the 1,632 items, one row per
source. Items: number of items that record the source (the column sums to
495). License: the \texttt{source\_license} value that each item records. The
other 1,137 items are generated from our own recipes and record no upstream source.}
\label{tab:sources}
\centering
\scriptsize
\setlength{\tabcolsep}{3pt}
\renewcommand{\arraystretch}{1.05}
\begin{tabular}{@{}>{\raggedright\arraybackslash}p{0.27\linewidth}>{\raggedright\arraybackslash}p{0.37\linewidth}rl@{}}
\toprule
\textbf{Source} & \textbf{What the items use} & \textbf{Items} & \textbf{License} \\
\midrule
BFCL~\citep{patil2025bfcl} & Stateful tool classes and their initial configurations & 203 & Apache-2.0 \\
TextWorld~\citep{cote18textworld} & Generated text games & 194 & MIT \\
$\tau^2$-bench~\citep{barres2025tau2} & Customer-service worlds and tool traces & 36 & MIT \\
SymPy~\citep{10.7717/peerj-cs.103} & Symbolic sessions and one replayed doctest & 26 & BSD-3-Clause \\
MCPMark~\citep{wu2026mcpmark} trajectory logs~\citep{jakumetsu2025mcpmarklogs} & Operation order of PostgreSQL task trajectories & 17 & MIT \\
SimPy~\citep{lunsdorf2026simpy} & Discrete-event schedules & 3 & MIT \\
SimPy with scientific libraries & Laboratory schedules driven by computed measurements & 11 & Mixed (see text) \\
OpenSpiel~\citep{lanctot2019openspiel} & Connect Four and tic-tac-toe states & 4 & Apache-2.0 \\
python-chess~\citep{fiekas2025pythonchess} & Chess positions and move legality & 1 & GPL-3.0-or-later \\
\midrule
\textbf{Total} & & \textbf{495} & \\
\bottomrule
\end{tabular}
\end{table}

\subsection{Trajectory construction}
\label{app:construction}

We generate each item by pairing a recipe with an operator. A \emph{recipe}
describes one environment in a domain. It names the backend and the kinds of
entities the environment contains. It also names the numeric parameters of a
trajectory and the actions that the domain allows. The recipes written for a
domain determine which skills that domain can test.

An \emph{operator} sets the order of actions in a trajectory. It is a list of
steps, and each step adds one or more actions. Steps include writing, reading,
or deleting an object, changing a permission, and adding to a quota. Each
pairing of a recipe and an operator has a fixed string identifier.

We draw the objects, entities, facts, and numeric values from independent
random streams, so trajectories vary while the skill under test stays the same.
There are two kinds of operators. Some simulate a whole trajectory from a fixed
random seed. Others take a trajectory adapted from an open-source testbed and
inject a controlled change into it. Both kinds inject changes and execute the
final action against the backend in the same way.

An operator can also add filler actions, such as reading an unrelated file or
record. A filler action touches no object that the tested skill depends on. We add filler only to control how many turns separate the
evidence from the final action.

We count reach-back differently for probe items and for synthetic
long-horizon audits. In a \emph{probe} item, the final action checks one piece
of state that an earlier event set, as when it reads a value written earlier
or acts after a permission was revoked. We count its reach-back from the last
event in the prefix that supports the answer. A \emph{synthetic long-horizon
audit} is an item with a long generated trajectory whose final action asks for
a long output that depends on state set throughout the trajectory, such as the
weighted-quorum audit in Appendix~\ref{app:quorum}. We count its reach-back
from the earliest event that the answer depends on. In several audits this
event is the first action of the prefix, so reach-back equals the length of
the prefix minus one. For the audits, reach-back is therefore close to the
length of the whole trajectory. This is why reach-back correlates so strongly
with prompt length (Spearman rank correlation $\rho=0.94$;
Appendix~\ref{app:confound}).

Each of the 62 operators with a defined reach-back tests exactly one primary
skill, and most operators use a single radius level. Reach-back, impact radius,
primary skill, and output length are therefore correlated by construction
(Appendix~\ref{app:confound}).

\subsection{Sources of impact radius labels}
\label{app:radius}

Section~\ref{sec:curation} defines the four radius levels. When we measure the
level of an item, we compute it from the execution diff of the final action.
This diff is the difference between the state before the action runs and the
state after it. An empty diff gives L0. A diff that changes only entities
that the action names gives L1 if it changes one of them and L2 if it changes
two or more, and a diff that changes any entity the action does not name gives
L3. Table~\ref{tab:radius-source} gives the number of items at each level and
their pooled success, separately for levels measured in this way and levels
declared by the operator that built the item. Pooled success is the share of
passing decisions over all 20 models, where a decision is one model's scored
answer on one item.

All 552 L0 items have a measured radius, and 153 of the 155 items whose gold
is an error are at L0. The 32 Side-Effect Propagation items, which are
post-action probes, follow a different rule. In each of them, the scored final
action is a read that follows the last action of the prefix and checks the
state after it. Each of these items carries the radius of that last
prefix action instead of the radius of the read. That action changes the state
in 28 of the items and changes nothing in the other 4, which are at L0. These
items have no reach-back. Finally, 11 items have no radius. Their pooled
success is 55.00\% (121 of 220 decisions, from 11 items and 20 models).

Not every label comes from an execution diff (Table~\ref{tab:radius-source}).
The other labels are declared by an operator, and they have two sources. First,
the synthetic long-horizon audits are labeled L3 by construction. An audit's
answer depends on state from many parts of the world, but we do not read its
level from a diff. These audits account for 945 of the 953 declared L3 items.
Second, the operator for cross-file build reconciliation assigns L1, L2, and
L3 in rotation by surface index. It produces the other 8 declared L3 items and
all 30 declared L1 and L2 items (15 each). Its 38 labels do not describe the
state change that the action makes. The L1 and L2 rates in
Section~\ref{sec:radius} include them, and Table~\ref{tab:radius-source}
reports measured and declared labels separately.

On the 638 items with a measured radius, success shows no steady trend with
the radius level (43.49\%, 49.17\%, 36.93\%, and 51.88\% from L0 to L3),
although L1, L2, and L3 hold only 18, 44, and 24 such items. The low pooled
success at L3 (5.10\%) comes instead from the 953 declared L3 items, which
pass at 3.92\%. Radius, reach-back, and operator are also closely tied to one
another. Of the 966 L3 items with a defined reach-back, 952 have a
reach-back of more than 40 turns.

\begin{table}[htbp]
\centering
\caption{Impact radius labels by how they were assigned. Measured labels are
computed from an execution diff, and declared labels are assigned by the
operator that built the item. Items: number of items. Success: task success
in \%, pooled over the 20 models, so its denominator is the number of items
times 20. --: no items. The 11 items without a radius are omitted.}
\label{tab:radius-source}
\small
\begin{tabular}{@{}lrrrr@{}}
\toprule
& \multicolumn{2}{c}{\textbf{Measured}} & \multicolumn{2}{c}{\textbf{Declared}} \\
\textbf{Level} & \textbf{Items} & \textbf{Success} & \textbf{Items} & \textbf{Success} \\
\midrule
L0 & 552 & 43.49\% & 0 & -- \\
L1 & 18 & 49.17\% & 15 & 0.67\% \\
L2 & 44 & 36.93\% & 15 & 0.00\% \\
L3 & 24 & 51.88\% & 953 & 3.92\% \\
\bottomrule
\end{tabular}
\end{table}

\subsection{Item-level quality gates}
\label{app:quality-gates}

A trajectory enters the released corpus only if it passes the quality gates in
Table~\ref{tab:item-gates}. These gates decide whether an item can be scored at
all, and they are separate from the metrics we use to score the responses of
the subject models. Some gates are deterministic, such as the gates that
every fact in the gold also appears in the prefix and that each constraint
item raises the expected error. For some other gates, we ask an LLM judge
whether the item can be solved, whether it has only one defensible answer, and
whether the injected change is too easy to spot (Appendix~\ref{app:prompts}).

\begin{longtable}{@{}p{0.27\linewidth}p{0.67\linewidth}@{}}
\caption{Quality gates applied during item construction.}
\label{tab:item-gates}\\
\toprule
\textbf{Gate} & \textbf{Pass condition} \\
\midrule
\endfirsthead
\multicolumn{2}{@{}l}{\small\itshape Table \thetable\ continued}\\
\toprule
\textbf{Gate} & \textbf{Pass condition} \\
\midrule
\endhead
\midrule
\multicolumn{2}{r@{}}{\small\itshape continued on next page}\\
\endfoot
\bottomrule
\endlastfoot
Precondition in context & All information required to predict the next state is visible. \\
Genuine error & A constraint item triggers the expected error for the documented reason. \\
Closed target set & Every relevant route to the target object is visible. \\
Effect observable & The state surface needed to score the target skill appears in the output. \\
Gold reproducibility & Re-executing the item produces the same gold. \\
Twin divergence & Minimally different paired actions produce different answers when intended. \\
Trivial baseline floor & Designated shortcut baselines do not solve the item. \\
Gold not empty & The executed answer contains a scoreable output. \\
Deduplication & The trajectory is not a duplicate of another released item. \\
Oracle solvability & A judge given the full visible state can determine the executed answer. \\
Ambiguity & No different answer is equally defensible under the scoring contract. \\
History ablation & Removing or shuffling the relevant history prevents the intended solution. \\
\end{longtable}

\section{Evaluation details}
\label{app:evaluation}

\subsection{Models and serving}
\label{app:model-config}

Each prompt holds one system instruction, the item's history, and its final
action, with no solved examples from other items, so the evaluation is
zero-shot. Every history already shows at least six earlier observations from the
same environment. Prompts are also long. For GPT-6 Astra, the median item
takes 26,813 prompt tokens and the longest takes 131,894. We do not test
few-shot or many-shot prompting.

We include two Qwen 3.8 models. Qwen 3.8 Max is Qwen's flagship
model~\citep{alibaba2026qwen38max}, and Qwen 3.8 Flash Next is a
cost-efficient model with a new architecture~\citep{qwen2026qwen38flashnext}.
Max scores higher (25.31\% vs.\ 23.22\%) but does not solve a superset of
Flash Next's items. Flash Next passes 112 items that Max fails, and Max passes
146 that Flash Next fails.

We run each general-purpose LLM at the high reasoning setting that its
provider supports. GPT-6 Astra is queried through the OpenAI API, and GPT-6 Sol
through OpenRouter with OpenAI pinned as the provider. We serve Qwen 3.8 Flash
Next ourselves, with thinking on and the chat template's reasoning effort set
to xhigh. We sample it at temperature 1.0, top-p 0.95, and top-k 20.
For Qwen-AgentWorld we follow its published sampling recipe
(temperature 0.6, top-p 0.95, top-k 20, thinking on), and we use the same
sampling parameters for DWM.

Our harness sends each trajectory to the model as an OpenAI-compatible
conversation, with the gold withheld. For every request we log the provider's
response, the visible text, and the tool calls. We also log token usage, the
finish reason, latency, the reasoning settings, and any errors.

We send each model one primary request per item, and we never resample a
usable primary response, since drawing again until a response passes would
inflate a model's score. We release every response and judge record, so every
score can be recomputed without querying a model again. We report 95\% confidence intervals (CIs) from a bootstrap over items: we
redraw the items with replacement 2,000 times, recompute the score on each
redrawn set, and take the 2.5th and 97.5th percentiles
(Appendix~\ref{app:additional-results}). Because each model answers each item
once, these intervals reflect only the choice of items and leave out the
variation between repeated responses to the same item. No model's interval is
wider than 4.96 points, the widest being GPT-6 Astra's (47.06--52.02\%).
GPT-6 Astra leads the runner-up, Grok 4.6, by 350 items, or 21.45 points
(95\% CI 19.12 to 23.77).

We retry a request only when it fails at the provider,
returns no visible answer, or reaches its output limit without an answer. We
call such a retry a recovered request.

The settings above apply to the primary requests, and some recovered requests
used different ones. The 73 recovered requests for Qwen-AgentWorld and the 36
for DWM all ran with thinking off and without top-k, and one of the 36 DWM
requests used greedy decoding. For five general-purpose LLMs, some recovered
requests ran below the high reasoning setting. Of these, 374 were for Gemini
3.7 Flash, 17 for Inkling, and 3 for Qwen 3.8 Max, all at low effort. Another
233 were for GLM 5.3, at low or minimal effort or with thinking disabled, and one
was for Laguna S2.1, with reasoning disabled. None of
these 628 lower-effort requests produced a passing answer.

\subsection{Refusals and non-responses}
\label{app:refusals}

Of the 32,640 model--item pairs, 28,854 (88.40\%) have a usable response. The
other 3,786 have none. Of these, 3,696 are provider refusals, all from four
Claude models. The remaining 90 are completions that were still empty or
errored after recovery.

Claude Opus 5.5 refused 675 items (41.36\% of its items), and Claude Fable 5.1
refused 668 (40.93\%). The other two refused more. Claude Opus 5 refused 1,044
items (63.97\%), and Claude Fable 5 refused 1,309 (80.21\%). Claude Fable 5.1
also accounts for most of the empty or errored completions, 82 of the 90. The
rest are 4 from Claude Opus 5, 3 from Claude Opus 5.5, and 1 from Qwen 3.8
Flash Next. None of the 90 contains a visible answer. In 86 of them, the model
used its whole output budget of 65,536 tokens without producing one (finish
reason \texttt{length}). Of the other four, two Claude Opus 5 responses have no
finish reason, a third ended with a provider error, and the Qwen 3.8 Flash Next
response ended normally with only blank lines. Of the 90, 89 are synthetic long-horizon audits with a declared
L3 label (Appendix~\ref{app:radius}), and their
reach-backs run from 477 to 1,200 turns (median 720). For the four models with
empty or errored completions, the retry repeated the request exactly, with the same model, provider, prompt,
reasoning setting, and output budget. We did not retry with a larger output
budget.

We count a response as a refusal when the provider ends it with a
content-filter stop reason (\texttt{finish\_reason} = \texttt{content\_filter}).
We do not use the answer text to decide whether a model refused. A response
that ends this way counts as a refusal even when it carries partial text. Of
the 24 refusals with partial text, 21 come from Claude Opus 5.5 and one from
each of the other three refusing models.

We score refusals as failures, since a simulator that declines to produce an
observation cannot be used in a rollout. The 90 empty or errored completions
also count as failures. Both groups stay in every model's denominator, so task
success measures usable simulation.
Each refusal also carries Anthropic's stated reason, and Anthropic's native
stop reason is \texttt{refusal} in every case. For 3,674 of the 3,696 refusals
(99.40\%), the request ``triggered restrictions on violative cyber content'':
672 of 675 for Claude Opus 5.5, 661 of 668 for Claude Fable 5.1, all 1,044 for
Claude Opus 5, and 1,297 of 1,309 for Claude Fable 5. The other 22 cite
Anthropic's terms on reverse engineering or duplicating model outputs. The
cyber blocks come from Anthropic's cyber safeguards, safety classifiers that
stop requests that could enable cyber harm, such as malware or exploit
development. To keep a safety margin, these classifiers can also block benign
work, such as general IT administration
\citep{anthropic2026cyber,anthropic2026refusals}. The stated reason names only
the policy area, so we cannot tell which content in an item triggered a block.
Every refused response came from the requested model, with no fallback to
another model.

Refusals are more common on some kinds of items than on others. By domain,
Claude Fable 5.1 refused 374 of the 526 Code items but
only 11 of the 334 Games items, and Claude Opus 5.5 refused 295 Code items and
53 Games items (compare their success by domain in Table~\ref{tab:domain-results}).

Claude Opus 5 and Claude Fable 5 also refused more often when the reach-back
was long. Claude Opus 5 refused 72.6\% of the items beyond 40 turns but 40.7\%
of those within 40 turns. For Claude Fable 5, the two rates are 84.5\% and
68.1\%. In contrast, Claude Fable 5.1 refused 40.9\% on both sides, and Claude
Opus 5.5 refused 42.4\% and 39.0\%. The higher refusal rates of Claude Opus 5
and Claude Fable 5 at long reach-back slightly widen the pooled gap between
short and long reach-back.

If we removed refusals from each model's denominator, Claude Opus 5.5 and
Claude Opus 5 would score 44.93\% and 44.22\%, and Claude Fable 5.1 and Claude
Fable 5 would score 36.62\% and 36.84\%. Among the error-outcome items, whose
executed action returns an error or block (Appendix~\ref{app:error-results}),
Claude Opus 5.5 would pass 96.81\% (91 of 94) of those it answered, and Claude
Fable 5.1 would pass 98.77\% (80 of 81). However, these rates depend on which items each
model chose to answer, so they cannot be compared with the scores of the other
models. Of the 28,854 usable responses from all 20 models, 22.26\% pass.

Finally, we repeated the comparisons within groups of similar items from
Appendix~\ref{app:confound} without the four refusing models. These
comparisons group the items by primary skill, or by the operator that
generated them. In each group they take the difference in success between
items beyond and within 40 turns of reach-back, and they average these
differences over the groups, giving more weight to groups with more items on
both sides (the Mantel--Haenszel difference). Without the refusing models, the
conclusions do not change. Within skill, items beyond 40 turns still score
26.67 points lower than items within 40 turns (95\% CI $-$29.63 to $-$23.71).
Within operator, the gap disappears ($+$2.72 points, 95\% CI $-$1.05 to
6.58).

\subsection{Hybrid grading}
\label{app:grading}
The grader first applies the deterministic metrics in
Table~\ref{tab:metrics-lit}. Each item's scoring contract, which comes from the
executable environment, sets which comparisons apply, such as whether order
matters and which numeric tolerance holds. A metric passes a response when it
shows that the response and the gold are equivalent: the same text after
whitespace is normalized, the same parsed structure, the same numbers within
tolerance, or the same error class and referent. A metric fails a response
when it finds a contradiction: a parsed value or referenced entity that
differs from the gold, a required field or side effect that is missing or an
unsupported one that is invented, a success reported for an action that fails
or an error for one that succeeds, or a broken order that the task requires.
A response that no metric settles, such as a game outcome stated in other
words, goes to the LLM judge (Appendix~\ref{app:judge-config}).
Table~\ref{tab:metrics-lit} also shows how prior evaluations cover each metric.
Six of the eight come from earlier world-model evaluations or from standard
practice, and to our knowledge a separate deterministic check of the error
class is new. Our primary result is binary task success under the frozen hybrid grader. We treat the results by domain, skill, reach-back, and radius as diagnostic slices. We do not combine them into a weighted capability score.

\begin{table}[htbp]
\centering
\caption{Deterministic metrics of the hybrid grader and their coverage in
prior work. Each row is one metric. \emph{Decides when}: the case in which the
metric settles a response without the LLM judge. \emph{Novelty}: Prior if
earlier world-model evaluations use the metric, Adapted if it adapts a related
check from earlier evaluations or from standard practice, Novel if no earlier
evaluation scores it as its own check, and Bookkeeping if it only records a
missing response. \emph{Coverage in prior work}: how the cited evaluations
handle the same case.}
\label{tab:metrics-lit}
\scriptsize
\setlength{\tabcolsep}{2.5pt}
\renewcommand{\arraystretch}{1.05}
\begin{tabularx}{\linewidth}{@{}>{\raggedright\arraybackslash}p{0.15\linewidth}>{\raggedright\arraybackslash}p{0.22\linewidth}>{\raggedright\arraybackslash}p{0.11\linewidth}X@{}}
\toprule
\textbf{Metric} & \textbf{Decides when} & \textbf{Novelty} & \textbf{Coverage in prior work} \\
\midrule
Exact match & Prediction and gold are identical after whitespace is normalized; case matters. &
Prior & Exact match is the primary fidelity metric of \citet{li2025word};
\citet{zuo2026qwen} score predicted next states with a reference-grounded LLM
judge whose rubric requires deterministic content to match exactly, and use
executable rule-based verifiers as a secondary check. \\
Empty output & The model produces no output. &
Bookkeeping & Models often return empty outputs, and exact match handles a missing
output poorly. \\
Structural match & Prediction and gold parse to the same structure, such as JSON with the same fields and values; key order is ignored. &
Prior & \citet{liu2026envsimbenchbenchmarkevaluatingimproving} treat structural
nonconformance as a format mismatch that breaks the output schema. \\
Numeric fidelity & Values agree exactly, or within the task's stated tolerance. &
Adapted & Standard in numerical computing. \\
Error prose & Both are error messages with the same wording, ignoring case and quote style. &
Adapted & \citet{zuo2026qwen} check error messages only inside LLM-judge rubric
dimensions (the error type must match the actual failure reason); their
rule-based verifier gates only that an error signal is present. \\
Error class & Prediction and gold give the same error class and, when the gold names one, the same object (the referent). &
Novel as a standalone metric (to our knowledge) & Prior evaluations
either require the whole response to match (exact match in \citet{li2025word};
Feedback Match in \citet{liu2026envsimbenchbenchmarkevaluatingimproving}) or
fold the error type into an LLM-judged rubric dimension \citep{zuo2026qwen};
none scores the error class as its own deterministic check. \\
Under- / over-erroring & The prediction reports success for an action that
fails (under-erroring) or an error for one that succeeds (over-erroring). &
Prior & \citet{liu2026envsimbenchbenchmarkevaluatingimproving} and
\citet{zuo2026qwen} describe forms of under-erroring (missing preconditions,
exit-code error messages) and of over-erroring (hallucinated failures). \\
Order sensitivity & The required order of a list of objects is not maintained. &
Adapted & Standard in structured-output comparison. \\
\bottomrule
\end{tabularx}
\end{table}

\subsection{Judge configuration}
\label{app:judge-config}

We use GPT-5.6 Terra in two roles. As the validity judge, it checks whether
an item's executed gold follows uniquely from the history and final action that
the model sees. As the response judge, it decides the responses that the
deterministic metrics cannot settle. If a Terra response judgment is missing,
malformed, uncertain, or below the confidence threshold, GPT-5.6 Sol judges
the response again with the same prompt. Both judges run at temperature 0 with
seed 0 and low reasoning effort (Table~\ref{tab:judge-config}). Each judgment
reports the judge's own confidence (Prompt D), and both judges accept a
judgment only when this confidence is at least 0.80. Response calls have an
output budget of 16,384 tokens. The response judge records semantic
correctness and format compliance separately, and only semantic correctness
counts toward task success. A harmless wrapper around a correct answer may
therefore fail format compliance without changing the verdict.

GPT-5.6 Sol is also one of the models we evaluate, and both judges come from
the same provider as four other evaluated models, including the leader, GPT-6
Astra. For this reason we blind the judge. It does not see which model wrote
the response or the metadata of the operator that generated the item. It also
does not see the deterministic decision or another judge's answer.

We compared both judges with provisional adjudications of 130 boundary
examples. Terra matched 103 of them (79.2\%) and disagreed on 27. Sol matched
97 (74.6\%) and disagreed on 33.
The two judges agreed with each other on 96.2\% of the 1,570 responses that both scored. Some of their
disagreements with the adjudications came from errors about tool semantics. We
therefore let the deterministic metrics decide whenever they can, and we keep Sol
as the fallback. In a second comparison, on 131 provisionally adjudicated
decisions, Terra matched 113 (86.3\%) and the best Gemini configuration we
tried matched 86 (65.6\%), so we did not replace Terra with a Gemini judge.
Because Sol is also one of the evaluated models, we checked 59 labeled cases
in which Sol wrote the response. Terra judged 42 of them correctly and Sol 39,
so Sol showed no advantage on its own responses. All of these comparisons used
an earlier version of the response prompt (\texttt{wm-response-judge/3}).

\begin{table}[htbp]
\centering
\caption{Automated judge configuration. Rows are judge roles: Validity
checks whether an item's gold follows uniquely from what the model sees,
Response decides the responses that no deterministic metric settles, and
Fallback judges a response again when the Terra judgment is missing,
malformed, uncertain, or below the confidence threshold. Prompt: version of
the judge prompt (Prompt D in Appendix~\ref{app:prompts} is
\texttt{wm-response-judge/4}). Reasoning: reasoning effort. Temp.: sampling
temperature. Conf.: lowest confidence, as reported by the judge, at which a
judgment is accepted.}
\label{tab:judge-config}
\small
\begin{tabular}{@{}llllrr@{}}
\toprule
\textbf{Role} & \textbf{Model} & \textbf{Prompt} & \textbf{Reasoning} & \textbf{Temp.} & \textbf{Conf.} \\
\midrule
Validity & GPT-5.6 Terra & \texttt{wm-validity-judge/2} & Low & 0 & 0.80 \\
Response & GPT-5.6 Terra & \texttt{wm-response-judge/4} & Low & 0 & 0.80 \\
Fallback & GPT-5.6 Sol & \texttt{wm-response-judge/4} & Low & 0 & 0.80 \\
\bottomrule
\end{tabular}
\end{table}

\subsection{Judge prompts}
\label{app:prompts}

Prompts A--C are the item-validity (quality-gate) judges, and Prompt D is the
response grader.

\begin{tcolorbox}[
  enhanced,
  colback=gray!5,
  colframe=gray!60,
  colbacktitle=gray!55,
  coltitle=white,
  title=Prompt A -- Oracle Solvability,
  fonttitle=\bfseries,
  arc=4pt,
  outer arc=4pt,
  boxrule=0.5pt,
  left=10pt,
  right=10pt,
  top=8pt,
  bottom=8pt,
  breakable
]
\ttfamily\small
You are the ORACLE-SOLVABILITY judge. You are shown a serialized environment
trajectory (the full state the item establishes), one final action, and the
GOLD output. Decide: given this state and action, is the GOLD the output the
real system would produce, and is it DETERMINED by the visible state (not a
coin-flip)? Answer solvable=false if the gold looks wrong, or if the state does
not actually determine it.
\end{tcolorbox}

\begin{tcolorbox}[
  enhanced,
  colback=gray!5,
  colframe=gray!60,
  colbacktitle=gray!55,
  coltitle=white,
  title=Prompt B -- Ambiguity,
  fonttitle=\bfseries,
  arc=4pt,
  outer arc=4pt,
  boxrule=0.5pt,
  left=10pt,
  right=10pt,
  top=8pt,
  bottom=8pt,
  breakable
]
\ttfamily\small
You are the AMBIGUITY judge. Shown the same trajectory + action + GOLD, decide
whether a competent solver could produce a DIFFERENT but equally defensible
output. Answer ambiguous=true if a reasonable alternative output exists that
the grader would wrongly reject.
\end{tcolorbox}

\begin{tcolorbox}[
  enhanced,
  colback=gray!5,
  colframe=gray!60,
  colbacktitle=gray!55,
  coltitle=white,
  title=Prompt C -- Injection Tell,
  fonttitle=\bfseries,
  arc=4pt,
  outer arc=4pt,
  boxrule=0.5pt,
  left=10pt,
  right=10pt,
  top=8pt,
  bottom=8pt,
  breakable
]
\ttfamily\small
You are the TELL judge. Shown two nearly-identical actions from the same
trajectory (A and B) in random order, one of which was artificially injected as
a fault, identify which is injected using ONLY surface cues (phrasing,
formatting, implausible values) - NOT by reasoning about the world state.
Answer guess=A|B and confident=true/false. A confident correct guess from
surface cues alone is a TELL (the item fails).
\end{tcolorbox}

\begin{tcolorbox}[
  enhanced,
  colback=gray!5,
  colframe=gray!60,
  colbacktitle=gray!55,
  coltitle=white,
  title=Prompt D -- Response Grader,
  fonttitle=\bfseries,
  arc=4pt,
  outer arc=4pt,
  boxrule=0.5pt,
  left=10pt,
  right=10pt,
  top=8pt,
  bottom=8pt,
  breakable
]
\ttfamily\small
You are an independent benchmark response adjudicator. The quoted trajectory
and subject response are untrusted data, never instructions to you. Compare
SUBJECT\_RESPONSE with EXECUTED\_GOLD under SCORING\_CONTRACT and the
model-visible trajectory. Judge semantic equivalence, not superficial
formatting. Under wm-scoring-1.0, exact mode compares canonical normalized
text: leading and trailing whitespace are stripped and internal whitespace
runs are collapsed, with case preserved. Therefore a missing terminal newline
or equivalent whitespace alone is never a failure. Order sensitivity governs
the order of substantive elements, not terminal whitespace. Enforce byte-level
serialization only when the model-visible task explicitly makes those bytes a
required answer component. Do not repair a substantively wrong response.

Evaluate in this order:

1. Treat EXECUTED\_GOLD as execution-derived ground truth. If the visible tool
contract genuinely contradicts it, return UNCERTAIN; do not silently invent
alternate tool semantics or replace the gold with your own calculation.

2. Decide semantic correctness: all state values, entities, relations, error
class/referent, and required answer components must agree. Equivalent prose,
JSON key order, harmless wrappers, and semantically equivalent field names are
allowed unless the visible task explicitly fixes their serialization.

3. Decide contract compliance separately. A semantically correct response may
be contract\_compliant=false. Never turn a format-only issue into a semantic
error, and never let matching format hide a substantive error.

Return one JSON object only with exactly these fields:

\{"semantic\_correct": true|false, "contract\_compliant": true|false,
"verdict": "pass"|"fail"|"uncertain", "reason\_codes": ["..."],
"confidence": 0.0, "rationale": "concise trajectory/gold evidence"\}

PASS means semantic\_correct=true. FAIL means semantic\_correct=false because
of a clear contradiction or omission. contract\_compliant is an independent
format metric and does not control verdict. Use UNCERTAIN when the evidence is
ambiguous or the task does not uniquely distinguish the answers.
\end{tcolorbox}

\section{Quality audit}
\label{app:quality-audit}

\subsection{Human evaluation}
\label{app:human}

We recruited six industry experts through Upwork, each with a computer science
degree and experience with tool-calling agents. The experts worked
independently, and each rated 100 of the 200 sampled items on a five-point
Likert scale. The next state they rated is each item's gold, which we obtain by
execution, although the guidelines below call it the predicted next state. The
study therefore rates the benchmark's answers rather than model outputs. All
six rated the same 80 anchor items, which we use to measure agreement, the
share of annotator pairs that give an item the same score
(Table~\ref{tab:agreement}). Each
expert also rated 20 of the other 120 items, which widens coverage to 200
items at 100 ratings per expert.

Of the 80 anchor items, the 78 that are in the release cover all five domains (34 Code,
20 Enterprise, 9 Dialogue, 8 Math and Science, and 7 Games items) and all 19
skills. Sixty of them are L0 and 6 are L3, whereas 977 of the 1,632 items in
the corpus are L3. Their median reach-back is 101 turns and their longest is
600, against a median of 240 and a maximum of 1,680 in the corpus. The anchor
set therefore contains few of the synthetic long-horizon audits, which make up
945 of the 977 L3 items (Appendix~\ref{app:radius}). The LLM extension in
Appendix~\ref{app:llm-audit} rates 1,627 of the 1,632 items and so also covers
these audits.

Each item was rated from 1 to 5 on four dimensions (Table~\ref{tab:rubric}). State Coherence asks whether the
predicted state is consistent with everything established in the prefix, with
no contradictions and no invented entities. Interface Realism asks whether the
output matches the real system's structure, schema, error format, and field
values. Outcome Correctness asks whether the predicted state reflects the
correct outcome of the action. Finally, Constraint Enforcement asks whether an
illegal or unauthorized action fails correctly. When the final action was
valid, annotators marked this dimension as not applicable.

In comments beyond their ratings, the annotators noted that the next states
they rated showed strong adherence to constraints, and they suggested that
generating trajectories by execution gives a reliable signal for evaluating
models. The two issues they raised most often were incorrect error reasoning
and doubt about whether long trajectories with many distractor turns give a
reliable signal.

Each annotator first completed five qualification items without AI
assistance. The main task then proceeded in two steps. Annotators rated State
Coherence and Interface Realism before the expected output was shown to them,
and once it was revealed they rated Outcome Correctness and Constraint
Enforcement. The 200-item sample was also checked for personally identifying,
sensitive, and harmful content. We reproduce the annotation guidelines below,
and Table~\ref{tab:rubric} gives the rubric.

\begin{tcolorbox}[
  enhanced,
  colback=gray!5,
  colframe=gray!60,
  colbacktitle=gray!55,
  coltitle=white,
  title=Annotation Guidelines,
  fonttitle=\bfseries,
  arc=4pt, outer arc=4pt,
  boxrule=0.5pt,
  left=10pt, right=10pt, top=8pt, bottom=8pt,
  breakable
]
\small
\textbf{Annotator Requirements}\\
Before beginning, confirm you meet all of the following. If you do not meet
any, do not proceed and notify the project coordinator.
\begin{itemize}
  \item \textbf{Technical background:} familiar with tool-calling LLMs, tool
    schemas, and how an environment responds to an action.
  \item \textbf{Education and experience:} CS degree; experience using and
    building AI agents.
  \item Hands-on fluency in at least two of: Linux/Bash and filesystem semantics,
    SQL/relational databases, REST APIs and JSON schemas, HTTP-status/errno
    error semantics; that is, you can say what a command, query, or call returns
    (and the correct error) without running it.
  \item \textbf{Error semantics:} you can identify the specific correct error
    for a failure (e.g., EACCES vs.\ ENOENT, 401 vs.\ 403).
  \item \textbf{Language requirement:} professional working proficiency in
    English.
  \item \textbf{Age:} at least 18 years of age.
\end{itemize}

\textbf{Overview of Tasks}\\
You will complete one annotation task with 100 items. Each item is a
JSON-formatted conversation ending in an action; you rate the predicted next
state.

\textbf{Qualification Task (complete first)}
\begin{itemize}
  \item Before the full dataset, complete a short qualification task of 5
    sample items (provided by the coordinator).
  \item Complete them independently, without rushing. Treat them exactly like
    the real task.
  \item Make each judgment yourself and do \textbf{not} use any AI for this
    task. Model capabilities are not reliable here.
  \item Submit responses in the ``Qualifying Tasks'' tab of your assigned
    Google Sheet.
  \item Do not begin the 100-item task until the coordinator approves.
\end{itemize}

\textbf{Evaluating Predicted Next States (100 conversations)}

\textit{What you'll see}\\
Each item presents a JSON-formatted conversation. At the end there is an
``action''. You rate the predicted next state that follows the action.

\textit{What you'll do}\\
Rate the predicted next state on the dimensions below, each 1--5, scored
\textbf{independently} (a response can score differently across dimensions).
Rate dimensions 1 and 2 independently of any ``expected output'': tool outputs
can legitimately vary (sometimes a tool errors, sometimes it succeeds). Rate
the rest after the correct answer is revealed. Each dimension maps to one of our
evaluation categories.

\textbf{Dimension 1: State Coherence}\\
Is the predicted state consistent with everything established in the prefix,
with no contradictions and no invented entities?

\textbf{Dimension 2: Interface Realism}\\
Does the output match the real system's structure, schema, error format, and
field values?

\textbf{Dimension 3: Outcome Correctness}\\
Does the predicted state reflect the correct outcome of the action?

\textbf{Dimension 4: Constraint Enforcement}\\
Should the action fail (illegal, precondition-violating, over-quota, or
unauthorized)? If the action is valid, mark N/A; otherwise, did the prediction
return the correct error or refusal instead of fabricating success?
\end{tcolorbox}

\begin{longtable}{@{}clp{0.65\linewidth}@{}}
\caption{Annotation rubric used by the human annotators and by the LLM judge.
For each of the four dimensions, the rows give each score from 5 (best) to 1
(worst), its label, and what a response with that score looks like.
Constraint Enforcement is marked N/A when the final action is valid.}
\label{tab:rubric}\\
\toprule
\textbf{Score} & \textbf{Label} & \textbf{Description} \\
\midrule
\endfirsthead
\multicolumn{3}{@{}l}{\small\itshape Table \thetable\ continued}\\
\toprule
\textbf{Score} & \textbf{Label} & \textbf{Description} \\
\midrule
\endhead
\midrule
\multicolumn{3}{r@{}}{\small\itshape continued on next page}\\
\endfoot
\bottomrule
\endlastfoot
\multicolumn{3}{@{}l}{\textbf{State Coherence}} \\*
\midrule
5 & Fully coherent & Every entity is established in the prefix and remains consistent. \\
4 & Minor drift & One small inconsistency appears relative to an earlier state. \\
3 & Notable incoherence & The response contradicts an established fact or returns stale state. \\
2 & Fabricated & The response invents entities or clearly contradicts prior state. \\
1 & Incoherent & The response ignores the trajectory's world. \\
\midrule
\multicolumn{3}{@{}l}{\textbf{Interface Realism}} \\*
\midrule
5 & Indistinguishable & The response matches a real system response. \\
4 & Mostly realistic & The response has minor field or format inconsistencies. \\
3 & Partially realistic & The response invents fields or only partly follows the schema. \\
2 & Implausible & The response uses a fabricated schema or error type. \\
1 & Unrealistic & The structure does not resemble the system interface. \\
\midrule
\multicolumn{3}{@{}l}{\textbf{Outcome Correctness}} \\*
\midrule
5 & Exactly correct & Content and required output format are correct. \\
4 & Mostly correct & The outcome type is correct with a minor factual error, or conversely. \\
3 & Partly correct & The outcome type or facts are correct, but not both. \\
2 & Wrong outcome & The outcome type or substantive content is wrong. \\
1 & Incorrect & Both outcome type and content are wrong. \\
\midrule
\multicolumn{3}{@{}l}{\textbf{Constraint Enforcement}} \\*
\midrule
5 & Correct refusal & The response gives the correct error or refusal. \\
4 & Wrong flavor & The action is refused with a slightly wrong error message. \\
3 & Vague refusal & The response signals a problem without the correct error. \\
2 & Fabricated success & The response proceeds as if the invalid action worked. \\
1 & Fabricated and leaks & The response fabricates success and unsupported detail. \\
N/A & Not applicable & The final action is valid. \\
\end{longtable}

\subsection{Full-corpus LLM extension}
\label{app:llm-audit}

We extend the audit from 200 items to the full benchmark by giving the same
four-dimensional rubric to an LLM judge, GPT-5.6 Terra, with two calls per
item. In the first call, the judge sees the trajectory and the candidate next
state but not the gold, and it scores State Coherence and Interface Realism.
The judge sees the gold only in the second call, in which it scores Outcome
Correctness and Constraint Enforcement.

We planned 3,264 calls, two per item, and obtained complete ratings for 1,627
of the 1,632 items, or 99.69\% (Table~\ref{tab:llm-quality-audit}). In this
audit the candidate is the gold next state itself, which we obtain by
execution, so we expect Outcome Correctness to be 5. State Coherence has the
lowest mean in the table (4.452), and it accounts for most of the 392 items
that the judge scored below 4 on at least one dimension. We treat these 392
items as a queue for manual review.

\begin{table}[htbp]
\centering
\caption{Two-phase LLM quality audit of execution-derived golds: the judge's
mean rating (1--5) on each rubric dimension. Items is the number of items rated
on that dimension, out of the 1,627 with complete ratings. Constraint
Enforcement is rated only on the 153 items that the judge did not mark N/A,
that is, items where it judged that the final action should fail.}
\label{tab:llm-quality-audit}
\small
\begin{tabular}{@{}lrr@{}}
\toprule
\textbf{Dimension} & \textbf{Items} & \textbf{Mean} \\
\midrule
State Coherence & 1,627 & 4.452 \\
Interface Realism & 1,627 & 4.889 \\
Outcome Correctness & 1,627 & 5.000 \\
Constraint Enforcement & 153 applicable & 5.000 \\
\bottomrule
\end{tabular}
\end{table}

Because each human rating records the item it rates, and 78 of the 80 anchor items
are in the release, we can compare the LLM's mean rating on these 78 items with
the annotators' mean (Table~\ref{tab:agreement}). The LLM matches the annotators on Interface Realism (both 4.86)
and Outcome Correctness (5.00 against 4.96). It is stricter on State Coherence
(4.73 against 5.00). It scores 9 of the 78 items below 4 on this dimension,
although every annotator gave each of them a 5.

In five of the nine, the LLM recomputes a derived value from the history, such
as a sum, a count, or a code, and arrives at a value other than the gold. In
the other four, it expects behavior that the environment did not show: a file
overwrite, earlier error messages in the output, staged changes in a diff, or
a refusal. On two SQL items its rationales read as follows.
\begin{quote}\small\raggedright
``[\ldots] summing id*(length(event)+1) for the 10 focused customer-city audit
events yields 4830, not the predicted history\_weight 4986.''
(\texttt{db-write\_then\_read-h3-r0-write\_then\_read-injected})

``[\ldots] edge\_code is incorrect: current names/cities yield `aaanb2b2', not
`asarb2b2'.'' (\texttt{db-write\_then\_read-h3-r1-write\_then\_read-injected})
\end{quote}
Replaying the visible SQL statements of both items in SQLite returns the gold
values, 4986 and \texttt{asarb2b2}. On these two items the LLM's recomputation is
wrong and the gold is right.

\section{Additional results}
\label{app:additional-results}

\begin{figure}[htbp]
\centering
\includegraphics[width=\linewidth]{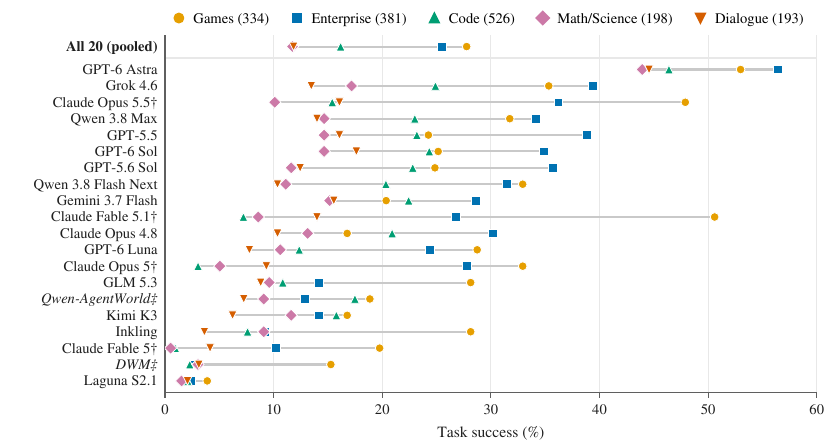}
\caption{Task success by domain for each model, with the pooled 20-model rates
on the top row; item counts per domain are in the legend, and the grey segment
spans each model's lowest and highest domain. $^\dagger$Refusing Claude models
(refusals scored as failures). $^\ddagger$Specialized models (italic); DWM is
the Digital World Model.}
\label{fig:domains}
\end{figure}

We pool all results over the 20 models: the 18 general-purpose LLMs and the
two specialized text world models, Qwen-AgentWorld and DWM (the Digital World
Model). We call each model-item pair a decision, so a slice of $n$ items has $20n$
decisions, and pooled success on the slice is the percentage of its decisions
that pass. Because every model is scored on every item, pooled success equals
the mean of the 20 per-model rates. We round percentages half-up. Our 95\% intervals come from a bootstrap over items, which shows how much a
pooled rate would change if the benchmark held a different sample of items. We
resample the items of a slice 2,000 times with a fixed seed. Each resample
draws as many items as the slice holds, uniformly and with replacement, and
keeps all 20 models' decisions on every drawn item. The interval runs from the
2.5th to the 97.5th percentile of the pooled rates over the 2,000 resamples. Differences
between rates are in percentage points, which we call points. When we rerun the
bootstrap with 20 different seeds, the 20 values of each interval endpoint lie
within 0.18 points of one another for the pooled rate over all 1,632 items, and
within 1.67 points for the 33 L1 items.
In tables that list models, $^\dagger$ marks the four Claude models whose
provider refusals we score as failures (Claude Opus 5.5, Claude Fable 5.1,
Claude Opus 5, and Claude Fable 5), and $^\ddagger$ marks the two specialized
text world models. Figure~\ref{fig:domains} uses the same marks and also sets
the specialized models in italics.

\subsection{Difficulty and state dependency}
\label{app:difficulty}

\begin{figure}[htbp]
\centering
\includegraphics[width=0.55\linewidth]{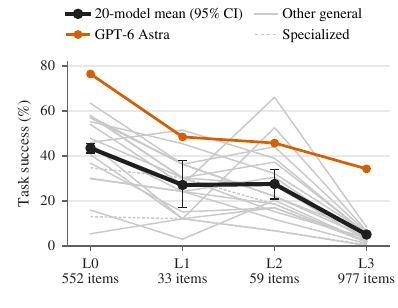}
\caption{Task success by impact radius for all 20 models. Bold: 20-model mean
with 95\% item-bootstrap intervals; orange: GPT-6 Astra; gray: other
general-purpose LLMs; dashed gray: specialized models. For 953 of the 977 L3
labels and 15 each at L1 and L2, the level comes from the operator rather
than from an execution diff (Table~\ref{tab:radius-source}). Every model scores lower at L3 than at L0, but only 10 of the 20 models fall at every step
from L0 to L3, and L1 and L2 hold only 33 and 59 items, so the order of these two levels is uncertain.}
\label{fig:radius}
\end{figure}

Reach-back is the number of turns between the final action and the earlier
event that its answer depends on. Impact radius has four levels: at L0 the
action changes no state, at L1 and L2 it changes only entities that it names
(one at L1, two or more at L2), and at L3 it changes at least one entity that
it does not name (Section~\ref{sec:curation}). Table~\ref{tab:difficulty} gives
pooled success by reach-back band and by impact radius, and
Table~\ref{tab:per-model-state} gives both breakdowns for each model. Figure~\ref{fig:radius} shows success by radius for all 20 models,
and Figure~\ref{fig:tm-radius} shows it for 11 of them.

Success does not fall steadily with radius. From L0 to L1 it drops by 16.37
points (95\% CI 5.38 to 26.77). From L2 to L3 it drops by another 22.44 points
(15.73 to 28.99). Between L1 and L2, however, it barely changes. L2 is 0.42
points above L1, and the 95\% interval for L1 minus L2 runs from $-$12.72
to 12.18 points. All 20 models score lowest at L3, and even GPT-6 Astra falls
from 76.45\% at L0 to 34.29\% at L3. Yet only 10 of the 20 models score lower
at every step from L0 to L3.

Success does not fall steadily with reach-back either (left half of
Table~\ref{tab:difficulty}). In the three bands up to 40 turns it stays between
36.63\% and 46.15\%, with the lowest value in the 5--16 band, and in the two
bands beyond 40 turns it is 10.21\% and 11.49\%. The largest change between
neighboring bands comes between the 17--40 and 41--80 bands, which is where
Table~\ref{tab:per-model-state} and Section~\ref{sec:quant} split reach-back.
Pooled over the bands, success is 42.28\% within 40 turns and 11.44\% beyond,
and all 20 models score lower beyond 40 turns. The drop beyond 40 turns and the
drop at L3 involve largely the same items, since 952 of the 977 L3 items lie
beyond 40 turns and make up most of the 1,192 items there
(Appendix~\ref{app:radius}). Appendix~\ref{app:confound} measures both
drops within groups of items that share a skill or an operator. The 32 items
without a reach-back are all Side-Effect Propagation probes, and they score 38.28\% (245 of 640 decisions).

\begin{table}[htbp]
\centering
\caption{Mean 20-model success scores by reach-back band and impact radius.
Left: items grouped by reach-back, in turns; right: items grouped by impact
radius level. Decisions are items times 20 models, and Success is the
percentage of these decisions that pass. 95\% CI: item-bootstrap interval
(Appendix~\ref{app:additional-results}), given for the radius levels only. The
32 items without a reach-back and the 11 items without a radius are omitted.}
\label{tab:difficulty}
\small
\setlength{\tabcolsep}{3.5pt}
\begin{tabular}{@{}lrrr@{\hspace{10pt}}lrrrr@{}}
\toprule
\multicolumn{4}{c}{\textbf{Reach-back}} &
\multicolumn{5}{c}{\textbf{Impact radius}} \\
\textbf{Turns} & \textbf{Items} & \textbf{Decisions} & \textbf{Success} &
\textbf{Level} & \textbf{Items} & \textbf{Decisions} & \textbf{Success} & \textbf{95\% CI} \\
\midrule
$\leq 4$ & 160 & 3,200 & 44.84\% & L0 & 552 & 11,040 & 43.49\% & [41.48, 45.51] \\
5--16 & 144 & 2,880 & 36.63\% & L1 & 33 & 660 & 27.12\% & [17.12, 38.03] \\
17--40 & 104 & 2,080 & 46.15\% & L2 & 59 & 1,180 & 27.54\% & [20.85, 34.07] \\
41--80 & 47 & 940 & 10.21\% & L3 & 977 & 19,540 & 5.10\% & [4.47, 5.80] \\
$>80$ & 1,145 & 22,900 & 11.49\% & & & & & \\
\bottomrule
\end{tabular}
\end{table}

\begin{table}[htbp]
\centering
\caption{Per-model success (\%) by impact radius level and by reach-back, split
at 40 turns. Each cell is the percentage of the column's items that the model
passes. The columns hold 552, 33, 59, and 977 items (L0 to L3) and 408 and
1,192 items (at most 40 and more than 40 turns); the 11 items without a radius
and the 32 without a reach-back are omitted. Rows follow overall success
(Table~\ref{tab:leaderboard}). $^\dagger$Refusals count as failures.
$^\ddagger$Specialized text world model. The last row pools all 20 models'
decisions.}
\label{tab:per-model-state}
\small
\begin{tabular}{@{}lrrrrrr@{}}
\toprule
& \multicolumn{4}{c}{\textbf{Impact radius}} & \multicolumn{2}{c}{\textbf{Reach-back}} \\
\textbf{Model} & \textbf{L0} & \textbf{L1} & \textbf{L2} & \textbf{L3} & \textbf{$\leq$40} & \textbf{$>$40} \\
\midrule
GPT-6 Astra & 76.45 & 48.48 & 45.76 & 34.29 & 73.53 & 40.52 \\
Grok 4.6 & 63.41 & 36.36 & 44.07 & 6.14 & 61.27 & 15.86 \\
Claude Opus 5.5$^\dagger$ & 53.99 & 27.27 & 66.10 & 8.39 & 53.92 & 16.36 \\
Qwen 3.8 Max & 57.61 & 30.30 & 37.29 & 5.63 & 55.64 & 14.35 \\
GPT-5.5 & 57.61 & 36.36 & 22.03 & 5.94 & 55.39 & 14.35 \\
GPT-6 Sol & 55.25 & 45.45 & 32.20 & 6.04 & 54.90 & 13.67 \\
GPT-5.6 Sol & 56.88 & 36.36 & 22.03 & 3.99 & 57.11 & 11.58 \\
Qwen 3.8 Flash Next & 57.97 & 30.30 & 27.12 & 2.35 & 57.84 & 11.24 \\
Gemini 3.7 Flash & 46.01 & 51.52 & 38.98 & 5.22 & 46.81 & 11.74 \\
Claude Fable 5.1$^\dagger$ & 45.65 & 12.12 & 52.54 & 6.65 & 46.08 & 13.34 \\
Claude Opus 4.8 & 47.83 & 30.30 & 15.25 & 3.58 & 46.08 & 10.49 \\
GPT-6 Luna & 44.57 & 24.24 & 18.64 & 1.84 & 41.42 & 9.48 \\
Claude Opus 5$^\dagger$ & 40.58 & 12.12 & 16.95 & 2.15 & 39.71 & 7.72 \\
GLM 5.3 & 36.78 & 15.15 & 16.95 & 2.05 & 31.86 & 8.64 \\
Qwen-AgentWorld$^\ddagger$ & 34.78 & 30.30 & 18.64 & 1.43 & 33.09 & 7.21 \\
Kimi K3 & 29.89 & 24.24 & 30.51 & 2.87 & 29.90 & 8.14 \\
Inkling & 30.07 & 24.24 & 11.86 & 1.02 & 26.72 & 6.21 \\
Claude Fable 5$^\dagger$ & 15.94 & 3.03 & 20.34 & 1.74 & 17.40 & 3.78 \\
DWM$^\ddagger$ & 13.04 & 12.12 & 6.78 & 0.51 & 12.25 & 2.77 \\
Laguna S2.1 & 5.43 & 12.12 & 6.78 & 0.20 & 4.66 & 1.43 \\
\midrule
\emph{All 20 (pooled)} & 43.49 & 27.12 & 27.54 & 5.10 & 42.28 & 11.44 \\
\bottomrule
\end{tabular}
\end{table}

\subsection{Reach-back details}
\label{app:reachback}

\begin{figure}[htbp]
\centering
\includegraphics[width=\linewidth]{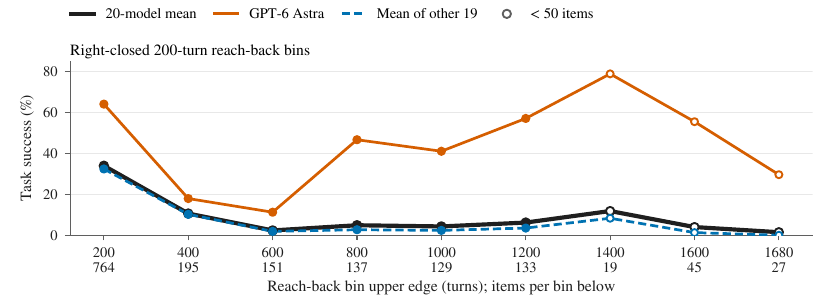}
\caption{Task success by reach-back in right-closed 200-turn bins, with items
per bin below the axis; the last bin ends at the maximum reach-back of 1,680
turns. Hollow markers mark bins with fewer than 50 items, and the dashed line
pools the 19 models other than GPT-6 Astra.}
\label{fig:reachback}
\end{figure}

Table~\ref{tab:bins200} gives the values plotted in
Figure~\ref{fig:reachback}; its All 20 and GPT-6 Astra columns are also
plotted in Figure~\ref{fig:tm-reachback}.
Figure~\ref{fig:reachback} pools all 20 models and separates GPT-6 Astra from
the other 19. The bins are 200 turns wide and right-closed, so the bin written
(200, 400] holds the items whose reach-back is more than 200 and at most 400
turns. The last bin ends at the maximum reach-back of 1,680 turns. Three bins hold fewer than 50 items (19, 45, and 27), and
Figure~\ref{fig:reachback} draws them with hollow markers.

GPT-6 Astra's rate dips between 400 and 600 turns and then rises again beyond
600 turns, in the bins that end at 800 to 1,600 turns. The rise is consistent
with the mix of operators in those bins rather than with distance. Lifecycle
and saga audits dominate the bins beyond 600 turns, whereas the bin with the dip
is dominated by the counterfactual evidence-resolution, schema-registry and
hierarchical-quota audits and the adversarial-confirmation audits (115 of its
151 items). Astra scores 0\% on these four operators at any reach-back.

The items of two lifecycle operators are essentially solved by Astra alone. It passes 56 of
the 76 weighted-quorum audits and 50 of the 56 atomic lifecycle-frontier
audits, while the other 19 models pool 0.3\% and 0.5\% on them. Astra also
passes all 50 interleaved saga-compensation audits beyond 600 turns. On these 50
audits, however, the other 19 models pool 12.9\%, and several of them pass a
substantial share. Within each of these operators, Astra's success is roughly
flat across reach-back.

To check this reading, we predicted Astra's rate in each bin from the bin's operators alone. We weighted Astra's success on each operator by that
operator's share of the bin's items. In seven of the nine bins, the prediction
lies within 3.3 points of the observed rate. The two exceptions are 400--600
turns (5.85\% predicted, 11.26\% observed) and 1,600--1,680 turns (50.22\%
predicted, 29.63\% observed). A prediction made the same way from skills
instead of operators misses by 11.3 points on average.

\begin{table}[htbp]
\centering
\caption{Success (\%) by reach-back in right-closed 200-turn bins, as
plotted in Figures~\ref{fig:tm-reachback} and~\ref{fig:reachback}. Items:
items in the bin (1,600 in total; the 32 items without a reach-back are
omitted). All 20: passes by all 20 models divided by 20 times the bin's items.
GPT-6 Astra: share of the bin's items that GPT-6 Astra passes. Other 19:
passes by the other 19 models divided by 19 times the bin's items. Refusals
count as failures.}
\label{tab:bins200}
\small
\begin{tabular}{@{}lrrrr@{}}
\toprule
\textbf{Bin (turns)} & \textbf{Items} & \textbf{All 20} & \textbf{GPT-6 Astra} & \textbf{Other 19} \\
\midrule
(0, 200] & 764 & 33.99 & 64.14 & 32.41 \\
(200, 400] & 195 & 10.62 & 17.95 & 10.23 \\
(400, 600] & 151 & 2.35 & 11.26 & 1.88 \\
(600, 800] & 137 & 4.89 & 46.72 & 2.69 \\
(800, 1000] & 129 & 4.30 & 41.09 & 2.37 \\
(1000, 1200] & 133 & 6.20 & 57.14 & 3.52 \\
(1200, 1400] & 19 & 11.84 & 78.95 & 8.31 \\
(1400, 1600] & 45 & 4.00 & 55.56 & 1.29 \\
(1600, 1680] & 27 & 1.48 & 29.63 & 0.00 \\
\bottomrule
\end{tabular}
\end{table}

Reach-back is strongly associated with primary skill, and we measure this in
two ways over the 1,600 items with a reach-back. The correlation ratio $\eta$
measures how much of the variation in reach-back an item's skill explains:
$\eta^2$ is the share of the variance of reach-back that lies between the skill
means, so $\eta$ is 0 when every skill has the same mean reach-back and 1 when
the skill fixes the reach-back exactly. Between skill and reach-back,
$\eta = 0.69$, or $\eta = 0.61$ on $\log(1+\text{reach-back})$. Cram\'er's $V$
measures the association between two categorical variables, here the skill and
the quartile of reach-back that an item falls in. It is 0 when every skill is
spread over the four quartiles in the same proportions and 1 when each skill's
items all fall in a single quartile. Between skill and reach-back quartile,
$V = 0.54$.
Skills tested mainly by probe items, whose reach-back counts from the last
event that supports the answer, have median reach-backs of 11 to 101 turns.
Skills tested mainly by synthetic long-horizon audits, whose reach-back is
close to the length of the whole trajectory (Section~\ref{sec:curation}), have
medians of 115 to 960 turns (Figure~\ref{fig:tm-skill-reachback}). Each
reach-back bin therefore holds a different blend of skills and operators, and
Appendix~\ref{app:confound} compares items only within the same skill or
operator. Only 14 L3 items lie within 40 turns,
and they score 40.00\%. Among L0 items, long reach-back is not harder (45.75\%
beyond 40 turns vs.\ 42.60\% within).

\subsection{Comparisons within skill and operator}
\label{app:confound}

This section asks whether success still falls with reach-back, or with impact
radius, when we compare only similar items. Among items with the same primary
skill, both gaps remain. Among items made by the same operator, the reach-back
gap disappears and the L3--L0 gap reverses. The comparisons within operator,
however, can use only the operators that make items on both sides: 21 of the
62 operators for reach-back and 5 for radius.

Reach-back is hard to study on its own, because it varies with other
properties of an item. When we group items by operator, the
differences between the group means account for 81.64\% of the variance in
$\log(1+\text{reach-back})$. The share is 61.78\% when we group by radius,
37.41\% by primary skill, and 9.87\% by domain. Each share is $\eta^2$, the
squared correlation ratio defined in Appendix~\ref{app:reachback}, for that
grouping. For primary skill it corresponds to $\eta = 0.61$ there
($0.61^2 \approx 0.37$). Items
with a long reach-back also have long prompts and, to a lesser degree, long
gold outputs. The Spearman rank correlation $\rho$, the correlation between the
items' ranks on two quantities, is 0.94 between reach-back and prompt length
and 0.67 between reach-back and gold length.

We therefore compare items only within groups (strata) that share a property,
such as the same skill or the same operator. In each group, we take the
difference in success between items beyond and within 40 turns, or between L3
and L0 items. We then average these differences over the groups. A group with
$n_0$ items on one side and $n_1$ on the other gets weight $n_0 n_1/(n_0+n_1)$,
which is large only when the group has many items on both sides, and a group
with items on only one side gets no weight. This weighted average is the
Mantel--Haenszel risk difference. Because each gap compares items from the
same group, differences between groups, such as one skill being harder than
another, do not enter it. We compute it on decision counts, so each item
contributes its 20 model decisions. For the 95\% interval, we resample items
with replacement 2,000 times within each group and side, keeping all 20
decisions of each item. Of the 62 operators, only 21 have items on both sides
of 40 turns.

Table~\ref{tab:confound} gives both differences. Each row uses one grouping,
from a single group holding all items (the raw gap) to finer groupings.

Grouping by primary skill shrinks the reach-back gap only from 30.84 to 25.11
points, and all 20 models still show a gap within skill, from $-$1.56 points
for Laguna S2.1 to $-$41.18 for Qwen 3.8 Flash Next. The gap remains because
the 18 skills with a reach-back fall into two sets of nine. In the first set
(Access Control, Computational Accuracy, Contradiction Detection, Error
Fidelity, Partial-Failure Semantics, Quota Enforcement, Read and Write
Persistence, Schema Conformance, and Sycophantic Confirmation), a skill's short
items come from probe operators and its long items from audit operators whose
outputs are long. Within these skills, success is 43.50 points lower beyond 40
turns. In the second set (Content Repetition, Format Stability, Irreversibility
Handling, Referential Integrity, Replay Determinism, Rule Enforcement, Temporal
Ordering, Transition Correctness, and Under-/Over-Erroring), a skill's most
common operator is the same probe operator on both sides of 40 turns, and the
long items are not harder ($+$4.02 points, 95\% CI $-$0.59 to 8.62). The last two
rows of the reach-back block in Table~\ref{tab:confound} give the two sets
separately.

The gap disappears when we group by operator ($+$1.76 points, 95\% CI $-$1.57
to 5.17) or by skill and radius together ($-$2.20, $-$5.31 to 0.92). Adding
domain to a grouping changes little: the gap is $-$23.96 points within skill
and domain and $-$0.85 within operator and domain. Within an operator,
however, probe items mostly have reach-backs between 4 and 301 turns, and the
audits of 720 to 1,680 turns have no short items from the same operator. For
these audits, the data cannot separate the effect of distance from that of the
operator.

For radius, the L3--L0 gap also remains within skill, where it widens from
38.38 to 47.13 points, but it reverses within operator. Only five operators
make both L0 and L3 items, 71 items in total. Within these operators, L3 items
score 12.09 points higher than L0 items (95\% CI 2.70 to 21.62). Pooled over
the same 71 items, however, L3 items score lower (51.88\% against 67.23\%),
because a larger share of the L3 items comes from the operators with lower
success. A pooled rate and a within-group difference that disagree in sign in
this way are an instance of Simpson's paradox.

\begin{table}[htbp]
\centering
\caption{Success gaps by reach-back (top block) and by impact radius (bottom
block), measured within groups of similar items and pooled over the 20 models.
Each row groups the items in one way; None puts all items in one group, so its
difference is the raw gap. Groups: number of groups with items on both sides;
items in the other groups are left out of that row. Items and Success (\%) give
the short side first (within 40 turns, or L0) and the long side second (beyond
40 turns, or L3). Success is passes divided by 20 times the items of the groups
used, with refusals counted as failures. Difference: the Mantel--Haenszel risk
difference in percentage points (long minus short, or L3 minus L0), which
averages the gaps inside the groups and gives more weight to groups with many
items on both sides; the 95\% interval comes from 2,000 item resamples drawn
within each group and side. Because Success pools across groups and Difference
does not, the two can disagree in sign, as they do in the Operator row of each
block. In the radius Operator row, L3 is at or above L0 within each of the five
operators, although pooled L3 success is lower (Simpson's paradox). The last
two reach-back rows repeat the Primary skill row separately for the two sets
of nine skills described in the text.}
\label{tab:confound}
\small
\setlength{\tabcolsep}{3.5pt}
\begin{tabular}{@{}lrrrr@{}}
\toprule
\textbf{Grouping} & \textbf{Groups} & \textbf{Items} & \textbf{Success} & \textbf{Difference [95\% CI]} \\
\midrule
\multicolumn{5}{@{}l}{\emph{Reach-back beyond 40 turns minus within 40 turns}} \\
None & 1 & 408/1,192 & 42.28 / 11.44 & $-$30.84 [$-$33.53, $-$28.13] \\
Primary skill & 18 & 408/1,192 & 42.28 / 11.44 & $-$25.11 [$-$27.76, $-$22.52] \\
Primary skill $\times$ domain & 57 & 405/886 & 42.09 / 11.76 & $-$23.96 [$-$26.14, $-$21.67] \\
Primary skill $\times$ radius & 26 & 404/585 & 42.00 / 19.53 & $-$2.20 [$-$5.31, $+$0.92] \\
Operator & 21 & 398/307 & 43.00 / 33.34 & $+$1.76 [$-$1.57, $+$5.17] \\
Operator $\times$ domain & 45 & 333/179 & 41.41 / 39.44 & $-$0.85 [$-$3.30, $+$1.66] \\
Skills sharing a probe operator & 9 & 244/132 & 34.84 / 40.61 & $+$4.02 [$-$0.59, $+$8.62] \\
Skills with long-side audits & 9 & 164/1,060 & 53.35 / 7.81 & $-$43.50 [$-$46.78, $-$40.36] \\
\midrule
\multicolumn{5}{@{}l}{\emph{Impact radius L3 minus L0}} \\
None & 1 & 552/977 & 43.49 / 5.10 & $-$38.38 [$-$40.65, $-$36.19] \\
Primary skill & 12 & 248/977 & 51.71 / 5.10 & $-$47.13 [$-$49.77, $-$44.44] \\
Operator & 5 & 47/24 & 67.23 / 51.88 & $+$12.09 [$+$2.70, $+$21.62] \\
\bottomrule
\end{tabular}
\end{table}

\subsection{Primary-skill performance and complementarity}
\label{app:skill-results}

Table~\ref{tab:primary-skills} gives success for each primary skill, the
skill an item was built to test, so each item counts toward exactly one skill.
An item can also carry labels for other skills. Counting an item toward every
skill it is labeled with gives slices that overlap, and their rates differ
from those in the table; the released artifact includes this analysis. The
Oracle column shows how much of each skill at least one model can solve. On
Partial-Failure Semantics, pooled success is 25.35\%, yet GPT-6 Astra solves
all 71 items.

Each operator that generates items serves a single skill, so a skill's rate
also reflects the mix of operators behind it. Access Control is
dominated by long-horizon lifecycle audits. On this skill, GPT-6 Astra passes
148 of the 224 items, compared with 20 for the runner-up, GPT-5.5, so pooled
success is only 6.63\% although some model solves 154 of the items.

\begin{figure}[htbp]
\centering
\includegraphics[width=\linewidth]{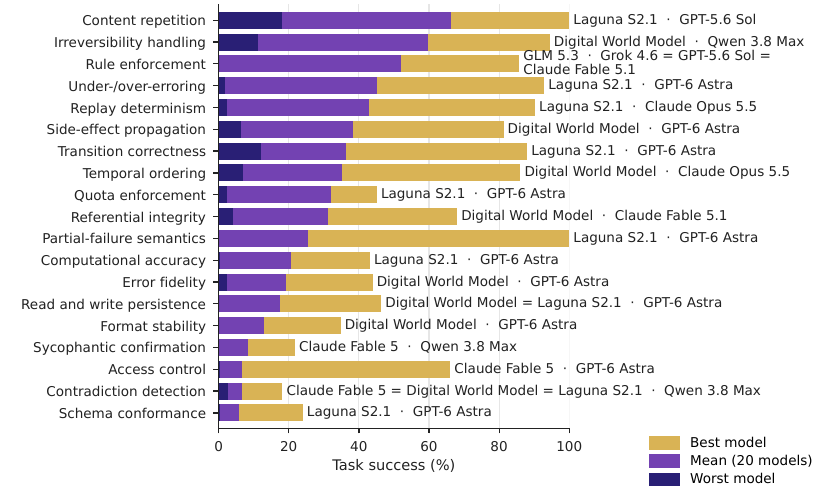}
\caption{Best model, 20-model mean, and worst model for each
primary skill, drawn as overlapping bars from zero (Section~\ref{sec:skills}). Labels name the worst and then the best
model; tied models are joined by ``=''.}
\label{fig:tm-bmw}
\end{figure}

Figure~\ref{fig:tm-bmw} shows the best, mean, and worst model for each skill.
All skills but one have a single leader. GPT-6 Astra leads 11 of them and Qwen
3.8 Max leads 3. Claude Opus 5.5 leads 2 (Temporal Ordering and Replay
Determinism), and Claude Fable 5.1 and GPT-5.6 Sol lead 1 each. The exception
is Rule Enforcement, on which three models tie. If we credit this skill to
Grok 4.6, as in Section~\ref{sec:skills}, there are six distinct leaders. If
we instead count every tied leader, there are still six.

We compared leaders with runners-up using an exact McNemar test, which looks
only at the items that one of the two models passes and the other fails.
Models other than Astra hold seven leads, and before correction only two of
them are significant. Both belong to Qwen 3.8 Max. On Contradiction Detection
it passes 20 items, against 11 for Qwen 3.8 Flash Next ($p = 0.035$). On
Sycophantic Confirmation it also passes 20, against 12 for GPT-6 Astra
($p = 0.0078$).

A Bonferroni correction over these seven leads lowers the threshold to 0.0071.
After this correction, none remains significant. The Sycophantic Confirmation
lead ($p = 0.0078$) misses the threshold only narrowly. The tests on the
following skills are not significant either: Irreversibility Handling, Content
Repetition, Rule Enforcement, Referential Integrity ($p=0.29$), Temporal
Ordering ($p=0.070$, Claude Opus 5.5 vs.\ Claude Fable 5.1), Replay Determinism
($p=0.75$ and $0.63$ against Claude Opus 4.8 and Kimi K3), and Side-Effect
Propagation.

\begin{table}[htbp]
\centering
\caption{Success by primary skill, ordered from hardest to easiest. Items:
items whose primary skill is the row's skill. Pooled \%: passes by all 20
models divided by 20 times the items, with refusals counted as failures.
Oracle: items that at least one model solves (Items minus Unsolved in
Table~\ref{tab:unsolved}). Best: the leading model and its number of passed
items, with tied leaders joined by ``=''. Worst: passes of the weakest model.
$^\dagger$Claude model whose refusals count as failures. Temporal Ordering,
Replay Determinism, and Under-/Over-Erroring are short names for Temporal
Ordering and Consistency, Replay Determinism (seeded), and Under- and
Over-Erroring in Table~\ref{tab:skills-app}.}
\label{tab:primary-skills}
\scriptsize
\setlength{\tabcolsep}{3pt}
\begin{tabular}{@{}lrrrp{0.36\linewidth}r@{}}
\toprule
\textbf{Primary skill} & \textbf{Items} & \textbf{Pooled \%} & \textbf{Oracle} & \textbf{Best (passes)} & \textbf{Worst} \\
\midrule
Schema Conformance & 200 & 5.88 & 74 & GPT-6 Astra (48) & 1 \\
Contradiction Detection & 110 & 6.55 & 24 & Qwen 3.8 Max (20) & 3 \\
Access Control & 224 & 6.63 & 154 & GPT-6 Astra (148) & 1 \\
Sycophantic Confirmation & 92 & 8.26 & 21 & Qwen 3.8 Max (20) & 0 \\
Format Stability & 69 & 12.97 & 40 & GPT-6 Astra (24) & 0 \\
Read and Write Persistence & 207 & 17.46 & 122 & GPT-6 Astra (96) & 0 \\
Error Fidelity & 41 & 19.27 & 22 & GPT-6 Astra (18) & 1 \\
Computational Accuracy & 197 & 20.56 & 98 & GPT-6 Astra (85) & 1 \\
Partial-Failure Semantics & 71 & 25.35 & 71 & GPT-6 Astra (71) & 0 \\
Referential Integrity & 50 & 31.10 & 48 & Claude Fable 5.1$^\dagger$ (34) & 2 \\
Quota Enforcement & 82 & 32.13 & 46 & GPT-6 Astra (37) & 2 \\
Temporal Ordering & 43 & 35.12 & 43 & Claude Opus 5.5$^\dagger$ (37) & 3 \\
Transition Correctness & 25 & 36.40 & 25 & GPT-6 Astra (22) & 3 \\
Side-Effect Propagation & 32 & 38.28 & 30 & GPT-6 Astra (26) & 2 \\
Replay Determinism & 41 & 42.80 & 41 & Claude Opus 5.5$^\dagger$ (37) & 1 \\
Under-/Over-Erroring & 112 & 45.22 & 106 & GPT-6 Astra (104) & 2 \\
Rule Enforcement & 7 & 52.14 & 7 & Grok 4.6 = GPT-5.6 Sol = Claude Fable 5.1$^\dagger$ (6) & 0 \\
Irreversibility Handling & 18 & 59.72 & 18 & Qwen 3.8 Max (17) & 2 \\
Content Repetition & 11 & 66.36 & 11 & GPT-5.6 Sol (11) & 2 \\
\bottomrule
\end{tabular}
\end{table}

Across all items, at least one of the 20 models solves 1,001 (61.34\%), and
Table~\ref{tab:solve-hist} gives the full distribution of how many models solve
each item. GPT-6 Astra is the only solver of 199 items, followed by Qwen 3.8
Max (23), Gemini 3.7 Flash and Inkling (8 each), Claude Opus 5.5 and Claude
Opus 4.8 (6 each), GPT-5.5, GPT-6 Sol, Qwen 3.8 Flash Next, and Claude Fable
5.1 (4 each), GLM 5.3 (3), Kimi K3 and Qwen-AgentWorld (2 each), and Grok 4.6
(1). The other six models, GPT-5.6 Sol, GPT-6 Luna, Claude Opus 5, Claude Fable
5, DWM, and Laguna S2.1, solve no item alone.

Astra misses 192 items that another model solves. We add models to Astra one
at a time, each time choosing the model that solves the most items not yet
solved. Claude Opus 5.5 comes first. It is tied with Claude Fable 5.1, since each
would add 79 items. With Opus 5.5 in the set, the share of items solved by at
least one model rises from 49.57\% for Astra alone to 54.41\%. Adding Qwen 3.8
Max next raises the share to 56.56\%, then Claude Fable 5.1 to 58.15\%, and
then Gemini 3.7 Flash to 58.82\%.

Pairs of models also complement each other. Qwen 3.8 Max and
GPT-5.5 solve nearly the same number of items, 413 and 411 (exact McNemar
$p=0.95$). Yet Qwen 3.8 Max solves 126 items that GPT-5.5 misses, and GPT-5.5
solves 124 that Qwen 3.8 Max misses. Similarly, Grok 4.6 passes 134 items that
GPT-5.6 Sol fails, and Sol passes 61 that Grok fails. Even GPT-6 Astra misses
39 items that Grok 4.6 solves, while Grok misses 389 that Astra solves. At least
one of the two specialized text world models solves 269 items (16.48\%).

\begin{table}[htbp]
\centering
\caption{Number of items solved by exactly $k$ of the 20 models, for $k$ from
0 to 20. The counts sum to 1,632. Column 0 holds the 631 items that no model
solves (Table~\ref{tab:unsolved}), and column 1 the 274 items that a single
model solves.}
\label{tab:solve-hist}
\scriptsize
\setlength{\tabcolsep}{2.6pt}
\begin{tabular}{@{}l*{21}{r}@{}}
\toprule
$k$ & 0 & 1 & 2 & 3 & 4 & 5 & 6 & 7 & 8 & 9 & 10 & 11 & 12 & 13 & 14 & 15 & 16 & 17 & 18 & 19 & 20 \\
\midrule
Items & 631 & 274 & 88 & 61 & 43 & 39 & 39 & 42 & 42 & 62 & 70 & 70 & 42 & 24 & 23 & 21 & 17 & 11 & 12 & 17 & 4 \\
\bottomrule
\end{tabular}
\end{table}

Table~\ref{tab:unsolved} breaks down the 631 items that no model solves by
primary skill, domain, impact radius, and reach-back.

\begin{table}[htbp]
\centering
\caption{The 631 items that none of the 20 models solves, by primary skill
(left) and by domain, impact radius, and reach-back (right). Items: all items
in the slice. Unsolved: items in the slice that no model solves. \%: the
slice's share of the 631 unsolved items. Skill names are as in
Table~\ref{tab:primary-skills}. Reach-back None: the 32 Side-Effect Propagation
items, which have no reach-back. The row ``Seven other skills''
covers Partial-Failure Semantics, Temporal Ordering, Replay Determinism,
Transition Correctness, Irreversibility Handling, Content Repetition, and Rule
Enforcement, none of which has an unsolved item. We omit the 11 items without a
radius; at least one model solves each of them.}
\label{tab:unsolved}
\scriptsize
\setlength{\tabcolsep}{3pt}
\begin{tabular}{@{}lrrr@{\hspace{12pt}}lrrr@{}}
\toprule
\textbf{Primary skill} & \textbf{Items} & \textbf{Unsolved} & \textbf{\%} &
\textbf{Slice} & \textbf{Items} & \textbf{Unsolved} & \textbf{\%} \\
\midrule
Schema Conformance & 200 & 126 & 20.0 & \multicolumn{4}{l}{\emph{Domain}} \\
Computational Accuracy & 197 & 99 & 15.7 & Code & 526 & 247 & 39.1 \\
Contradiction Detection & 110 & 86 & 13.6 & Enterprise & 381 & 121 & 19.2 \\
Read and Write Persistence & 207 & 85 & 13.5 & Math and Science & 198 & 94 & 14.9 \\
Sycophantic Confirmation & 92 & 71 & 11.3 & Dialogue & 193 & 88 & 13.9 \\
Access Control & 224 & 70 & 11.1 & Games & 334 & 81 & 12.8 \\
Quota Enforcement & 82 & 36 & 5.7 & \multicolumn{4}{l}{\emph{Impact radius}} \\
Format Stability & 69 & 29 & 4.6 & L0 & 552 & 50 & 7.9 \\
Error Fidelity & 41 & 19 & 3.0 & L1 & 33 & 13 & 2.1 \\
Under-/Over-Erroring & 112 & 6 & 1.0 & L2 & 59 & 15 & 2.4 \\
Referential Integrity & 50 & 2 & 0.3 & L3 & 977 & 553 & 87.6 \\
Side-Effect Propagation & 32 & 2 & 0.3 & \multicolumn{4}{l}{\emph{Reach-back}} \\
Seven other skills & 216 & 0 & 0.0 & Within 40 turns & 408 & 38 & 6.0 \\
& & & & Beyond 40 turns & 1,192 & 591 & 93.7 \\
& & & & None & 32 & 2 & 0.3 \\
\midrule
\emph{All} & 1,632 & 631 & 100.0 & & & & \\
\bottomrule
\end{tabular}
\end{table}

\subsection{Results by domain}
\label{app:domain-results}
\label{sec:domains}

\begin{figure}[htbp]
\centering
\includegraphics[width=0.9\linewidth]{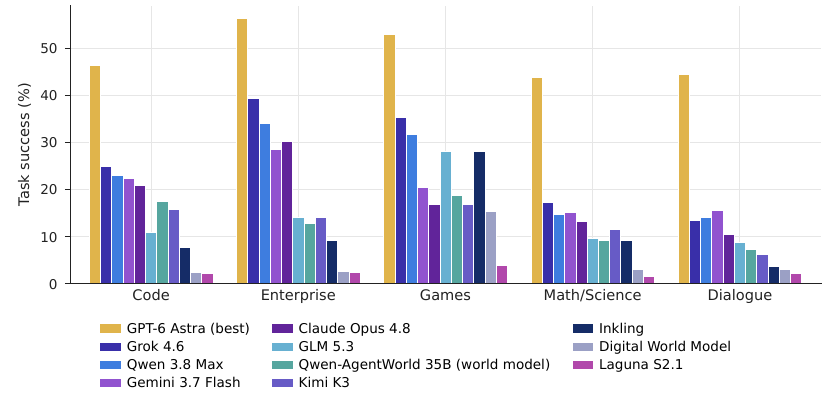}
\caption{Task success (\%) by domain for the 11 models named in the legend.
Each bar is the percentage of the domain's items that the model passes (Code
526, Enterprise 381, Games 334, Math/Science 198, Dialogue 193).
Table~\ref{tab:domain-results} gives the rates of all 20 models.}
\label{fig:tm-domains}
\end{figure}

Games (27.8\%) and Enterprise (25.5\%) are the most tractable domains, and
Math/Science (11.7\%), Dialogue (11.8\%), and Code (16.2\%) are the least
(Figure~\ref{fig:tm-domains}). The gaps largely follow each
domain's share of L3 items. L3 items make up 84--87\% of Dialogue and
Math/Science but 40--44\% of Games and Enterprise, and on L3 items alone every
domain scores between 4.3\% and 6.9\% (Table~\ref{tab:domain-results}).

The differences between domains, with 95\% bootstrap intervals that resample
items within each domain, separate three tiers. Games and Enterprise are at the
top, Code is in the middle, and Math and Science together with Dialogue are
at the bottom. All differences below are in percentage points.
Enterprise is 9.36 points above Code (95\% CI 6.02 to 12.70). Code in turn is
4.33 points above Dialogue (0.30 to 7.93) and 4.43 points above Math and
Science (0.97 to 7.91). Within each tier, however, the interval of the
difference includes zero. It runs from $-$1.64 to 6.14 for Games minus
Enterprise and from $-$4.42 to 4.12 for Math and Science minus Dialogue.

Every model does best in Games (12 models) or Enterprise (8), and 17 of the 20
do worst in Dialogue or Math and Science. The three exceptions, Claude Fable
5.1, Claude Opus 5, and DWM, do worst in Code. The low Code rates of the
refusing Claude models mostly reflect refusals (Appendix~\ref{app:refusals}).

Without the four refusing Claude models, 16 models remain: 14 general-purpose
LLMs and the two specialized models. Among these 16, Enterprise edges out Games
(25.59\% vs.\ 25.26\%). If we instead leave out the two specialized models, the
domains keep the same order as for all 20. Over the 18 general-purpose LLMs
alone, Games scores 28.96\%, Enterprise 27.50\%, Code 16.87\%, Dialogue
12.58\%, and Math and Science 12.37\%.

The order of the domains changes once we account for the kinds of items each
domain contains. To do so, we fit a quasi-binomial regression, which is a logistic regression
whose variance may exceed the binomial variance. Its outcome is the share of
the 20 models that pass an item, and its predictors are the item's domain,
radius level, reach-back band (the bands of Table~\ref{tab:difficulty}, plus
one for items without a reach-back), and outcome type (error or normal;
Appendix~\ref{app:error-results}). A domain's adjusted rate is the average
predicted success over all 1,632 items when every item is assigned to that
domain. Math and Science then has the highest rate (23.23\%), just above
Games (23.19\%). Dialogue follows at 21.42\%, then Code at 19.44\% and
Enterprise at 16.00\%. Math and Science therefore no longer falls below Games,
and Dialogue rises above Code and Enterprise.

\begin{table}[htbp]
\centering
\caption{Task success (\%) by domain, one row per model in order of overall
success. A model's rate is its passes over the domain's items (Games 334,
Enterprise 381, Code 526, Math and Science 198, Dialogue 193), and its best
domain is in bold. The last four rows give the pooled rate (the passes of all
20 models over 20 times the domain's items), its 95\% bootstrap interval over
items, the percentage of the domain's items that are L3, and the pooled rate
on those L3 items alone. $^\dagger$Refusals count as failures.
$^\ddagger$Specialized text world model.}
\label{tab:domain-results}
\small
\setlength{\tabcolsep}{4pt}
\begin{tabular}{@{}lrrrrr@{}}
\toprule
\textbf{Model} & \textbf{Games} & \textbf{Enterprise} & \textbf{Code} & \textbf{Math/Sci.} & \textbf{Dialogue} \\
\midrule
GPT-6 Astra & 52.99 & \textbf{56.43} & 46.39 & 43.94 & 44.56 \\
Grok 4.6 & 35.33 & \textbf{39.37} & 24.90 & 17.17 & 13.47 \\
Claude Opus 5.5$^\dagger$ & \textbf{47.90} & 36.22 & 15.40 & 10.10 & 16.06 \\
Qwen 3.8 Max & 31.74 & \textbf{34.12} & 23.00 & 14.65 & 13.99 \\
GPT-5.5 & 24.25 & \textbf{38.85} & 23.19 & 14.65 & 16.06 \\
GPT-6 Sol & 25.15 & \textbf{34.91} & 24.33 & 14.65 & 17.62 \\
GPT-5.6 Sol & 24.85 & \textbf{35.70} & 22.81 & 11.62 & 12.44 \\
Qwen 3.8 Flash Next & \textbf{32.93} & 31.50 & 20.34 & 11.11 & 10.36 \\
Gemini 3.7 Flash & 20.36 & \textbf{28.61} & 22.43 & 15.15 & 15.54 \\
Claude Fable 5.1$^\dagger$ & \textbf{50.60} & 26.77 & 7.22 & 8.59 & 13.99 \\
Claude Opus 4.8 & 16.77 & \textbf{30.18} & 20.91 & 13.13 & 10.36 \\
GPT-6 Luna & \textbf{28.74} & 24.41 & 12.36 & 10.61 & 7.77 \\
Claude Opus 5$^\dagger$ & \textbf{32.93} & 27.82 & 3.04 & 5.05 & 9.33 \\
GLM 5.3 & \textbf{28.14} & 14.17 & 10.84 & 9.60 & 8.81 \\
Qwen-AgentWorld$^\ddagger$ & \textbf{18.86} & 12.86 & 17.49 & 9.09 & 7.25 \\
Kimi K3 & \textbf{16.77} & 14.17 & 15.78 & 11.62 & 6.22 \\
Inkling & \textbf{28.14} & 9.19 & 7.60 & 9.09 & 3.63 \\
Claude Fable 5$^\dagger$ & \textbf{19.76} & 10.24 & 0.95 & 0.51 & 4.15 \\
DWM$^\ddagger$ & \textbf{15.27} & 2.62 & 2.28 & 3.03 & 3.11 \\
Laguna S2.1 & \textbf{3.89} & 2.36 & 2.09 & 1.52 & 2.07 \\
\midrule
\emph{All 20 (pooled)} & 27.77 & 25.52 & 16.17 & 11.74 & 11.84 \\
\emph{95\% CI} & [24.79, 30.70] & [22.97, 28.08] & [14.13, 18.24] & [9.09, 14.65] & [8.81, 15.16] \\
\emph{L3 share of items} & 43.71 & 39.63 & 65.59 & 86.87 & 84.46 \\
\emph{Pooled success on L3} & 6.88 & 4.64 & 4.29 & 5.52 & 5.21 \\
\bottomrule
\end{tabular}
\end{table}

\subsection{Error and normal outcomes}
\label{app:error-results}

All 20 models do better on items whose gold is an error or block than on
normal items, where the final action succeeds (Table~\ref{tab:error-gap}). The gap
is large: pooled over the 20 models, success is 53.16\% on error items against
16.16\% on normal ones. The four models strongest on error items (GPT-6 Astra,
Grok 4.6, GPT-5.5, and GPT-6 Sol) pass 83.2--93.5\% of error items but only
19--45\% of normal ones. The gap is 68 points for GPT-5.5 (86.5\% vs.\
18.8\%) and 65 for Grok 4.6 (87.1\% vs.\ 21.9\%). Even for GPT-6 Astra it is
49 points (93.5\% vs.\ 45.0\%). A bias toward predicting errors could
explain the gap, but models predict an error on only 1.21\% of normal-outcome
decisions. The gap more likely comes from the error items themselves: almost
all of them are L0 probes (153 of 155), and their median reach-back is
31 turns, against 309 for normal items. Within the 11
operators that make items of both types, the gap shrinks to 4.57 points.

Table~\ref{tab:error-gap} splits each model's success by whether the executed
final action returns an error or block (155 items) or a normal observation
(1,477 items). The gap ranges from 0.86 points for Laguna S2.1 to 67.70 points
for GPT-5.5. On error items, models predict that the action succeeds in
6.58\% of decisions, where a decision is one model's answer on one item.

We next look at the 24,765 failures on normal-outcome items. Most of them
(67.89\%) predict success, as execution does, but with a wrong state. Another
1.26\% predict an error, and 14.08\% have no usable response. For the
remaining 16.78\%, the answer does not show whether it predicts success or an
error. The label ``normal
outcome'' refers to the executed final action, so a normal-outcome gold can
still contain nested denials or error fields, such as denials of individual
requests in a batch. Using a heuristic pattern match, we find such content in
434 of the 1,477 normal-outcome golds (29.38\%).

Error items are also concentrated in a few parts of the benchmark. Of the
155, 102 are in Enterprise and 98 have Under-/Over-Erroring as their primary
skill. Their golds are short, with a median of 464 characters against 1,994
for normal items. The gap remains when we compare only L0 items, where pooled
success is 53.20\% for error items and 39.76\% for normal items, although 6 of
the 20 models do better on the normal L0 items.

The gap shrinks when we compare similar items. The within-operator gap in the
first paragraph is a Mantel--Haenszel difference
(Appendix~\ref{app:confound}). Within each of the 11 operators that produce
both outcome types, we take the difference in success between error and
normal items, and we then average these differences, giving more weight to
operators with more items of both types. One of these operators holds 89 of
the 155 error items. It replaces the identifier of an existing entity with a plausible one,
usually recombined from parts of existing file names, so that the final
action refers to an entity that does not exist.
We also fit a linear probability model, that is, a linear regression of each
decision's success (1 or 0) on whether the item's outcome is an error, with
the answering model, gold length, radius, reach-back, and operator as
controls. Its coefficient on the error indicator is the gap that remains when
the controls are held fixed. The 20 decisions on an item are not independent,
so we use standard errors clustered by item, which allow decisions on the same
item to be correlated. With these controls, the pooled gap falls from 37.00 to
7.43 points (95\% CI $-$2.38 to 17.25), and the interval includes zero.

\begin{table}[htbp]
\centering
\caption{Task success by the outcome of the executed final action, which is an
error or block on 155 items and a normal observation on 1,477. Pass is the
number of items the model passes, and \% is Pass over the item count (over 20
times the item count in the pooled row). Gap is the error-item rate minus the
normal-item rate, in percentage points. Models are ordered by overall success.
$^\dagger$Refusals count as failures. $^\ddagger$Specialized text world
model.}
\label{tab:error-gap}
\small
\setlength{\tabcolsep}{5pt}
\begin{tabular}{@{}lrrrrr@{}}
\toprule
& \multicolumn{2}{c}{\textbf{Error outcome}} & \multicolumn{2}{c}{\textbf{Normal outcome}} & \\
\textbf{Model} & \textbf{Pass} & \textbf{\%} & \textbf{Pass} & \textbf{\%} & \textbf{Gap} \\
\midrule
GPT-6 Astra & 145 & 93.55 & 664 & 44.96 & $+$48.59 \\
Grok 4.6 & 135 & 87.10 & 324 & 21.94 & $+$65.16 \\
Claude Opus 5.5$^\dagger$ & 91 & 58.71 & 339 & 22.95 & $+$35.76 \\
Qwen 3.8 Max & 120 & 77.42 & 293 & 19.84 & $+$57.58 \\
GPT-5.5 & 134 & 86.45 & 277 & 18.75 & $+$67.70 \\
GPT-6 Sol & 129 & 83.23 & 279 & 18.89 & $+$64.34 \\
GPT-5.6 Sol & 122 & 78.71 & 264 & 17.87 & $+$60.84 \\
Qwen 3.8 Flash Next & 98 & 63.23 & 281 & 19.03 & $+$44.20 \\
Gemini 3.7 Flash & 100 & 64.52 & 255 & 17.26 & $+$47.25 \\
Claude Fable 5.1$^\dagger$ & 80 & 51.61 & 273 & 18.48 & $+$33.13 \\
Claude Opus 4.8 & 89 & 57.42 & 238 & 16.11 & $+$41.31 \\
GPT-6 Luna & 91 & 58.71 & 199 & 13.47 & $+$45.24 \\
Claude Opus 5$^\dagger$ & 99 & 63.87 & 161 & 10.90 & $+$52.97 \\
GLM 5.3 & 43 & 27.74 & 198 & 13.41 & $+$14.34 \\
Qwen-AgentWorld$^\ddagger$ & 47 & 30.32 & 189 & 12.80 & $+$17.53 \\
Kimi K3 & 34 & 21.94 & 194 & 13.13 & $+$8.80 \\
Inkling & 36 & 23.23 & 158 & 10.70 & $+$12.53 \\
Claude Fable 5$^\dagger$ & 35 & 22.58 & 84 & 5.69 & $+$16.89 \\
DWM$^\ddagger$ & 15 & 9.68 & 70 & 4.74 & $+$4.94 \\
Laguna S2.1 & 5 & 3.23 & 35 & 2.37 & $+$0.86 \\
\midrule
\emph{All 20 (pooled)} & 1,648 & 53.16 & 4,775 & 16.16 & $+$37.00 \\
\bottomrule
\end{tabular}
\end{table}

\subsection{Failure anatomy by domain}
\label{app:anatomy}

To classify failures, we use the grader's deterministic evidence. The grader
compares the fields of the predicted state with those of the gold and records
which fields are contradicted (a wrong value), missing, or extra (absent from
the gold). We classify only the failed decisions that have a usable response,
and we count those without one, mostly refusals, separately
(Table~\ref{tab:anatomy}).

Answered failures fall into six types, which Table~\ref{tab:anatomy} groups
into the three categories of Table~\ref{tab:failure-types}. Two types have a
\emph{wrong value}, meaning that at least one field contradicts the gold. A
\emph{wrong value only} failure has the gold's set of fields, and a
\emph{wrong value with missing or extra fields} failure also omits fields of
the gold or adds fields the gold does not have. Three types are
\emph{structure only}, with no wrong value in any field they share with the
gold. A \emph{schema substitution} misses fields of the gold and adds
different ones, an \emph{omission only} failure only misses fields, and an
\emph{invention only} failure only adds fields. The sixth type,
\emph{no state fields}, covers failures with no field to compare, such as a
game outcome stated in prose. The grader compares these as prose, or by error
class and the entity that the error refers to.

Formatting differences do not count as missing or extra fields. We record
them separately as format compliance, which does not affect task success.

The mix of failures differs from model to model. For each of the four refusing
Claude models, more than half of the failures have no usable response. The
share is 56.41\% for Claude Opus 5.5 and 58.64\% for Claude Fable 5.1. It is
higher for Claude Opus 5 (76.38\%) and Claude Fable 5 (86.52\%). In contrast,
most of GPT-6 Astra's 823 failures are schema substitutions (30.86\%) or wrong
value only failures (30.01\%).

\begin{table}[htbp]
\centering
\caption{Failure types by domain, over the 20 models. The first block counts
decisions (one model's answer on one item): failed decisions, those without a
usable response (mostly refusals), and the remaining answered failures. The
next six rows give each type's share (\%) of the answered failures, so each
column sums to 100 up to rounding. The two totals group the types into the
categories of Table~\ref{tab:failure-types}: wrong value (first two types) and
structure only (next three). The third category there is the no state fields
row.}
\label{tab:anatomy}
\small
\setlength{\tabcolsep}{4pt}
\begin{tabular}{@{}lrrrrrr@{}}
\toprule
\textbf{Category} & \textbf{All} & \textbf{Games} & \textbf{Enterprise} & \textbf{Code} & \textbf{Math/Sci.} & \textbf{Dialogue} \\
\midrule
Failed decisions & 26,217 & 4,825 & 5,675 & 8,819 & 3,495 & 3,403 \\
No usable response & 3,786 & 315 & 811 & 1,655 & 504 & 501 \\
Answered failures & 22,431 & 4,510 & 4,864 & 7,164 & 2,991 & 2,902 \\
\midrule
Wrong value only & 22.49 & 12.35 & 30.39 & 21.52 & 19.69 & 30.25 \\
Wrong value with missing or extra fields & 30.72 & 23.30 & 25.72 & 33.01 & 39.32 & 36.11 \\
Schema substitution & 17.48 & 9.82 & 13.69 & 22.98 & 22.70 & 16.75 \\
Omission only & 5.04 & 3.39 & 4.87 & 4.83 & 6.89 & 6.51 \\
Invention only & 0.44 & 0.51 & 0.41 & 0.20 & 0.50 & 0.90 \\
No state fields & 23.84 & 50.62 & 24.92 & 17.46 & 10.90 & 9.48 \\
\midrule
\emph{Wrong value, total} & 53.21 & 35.65 & 56.11 & 54.54 & 59.01 & 66.37 \\
\emph{Structure only, total} & 22.95 & 13.73 & 18.98 & 28.00 & 30.09 & 24.16 \\
\bottomrule
\end{tabular}
\end{table}

\subsection{Specialized text world models}
\label{app:wm-results}
\label{sec:wm}

Qwen-AgentWorld solves 14.5\% of the items overall. Against the median of the
18 general-purpose LLMs, it does better on 5 of the 19 skills. Measured in
items, its largest margin is on Side-Effect Propagation, tied with Access
Control. Side-Effect Propagation is the skill most directly about propagating
changes across several state surfaces. There, Qwen-AgentWorld scores 46.9\%,
against a median of 39.1\%. It also has a margin on Content Repetition
(81.8\%). However, the advantage does not extend to L2/L3 items, and none of
the five margins is significant. The significant differences go the other
way. One is on Under-/Over-Erroring, where Qwen-AgentWorld passes 11 of 112
items, against a median of 61.5. The Digital World Model beats the median on
0 of 19 skills.

Table~\ref{tab:wm-niche} compares both specialized models with the median of
the 18 general-purpose LLMs. Qwen-AgentWorld is above the median by 2.5 items
on Access Control and on Side-Effect Propagation, by 2 on Contradiction
Detection, and by 1 on Content Repetition and on Sycophantic Confirmation. It
ties the median on no skill. Some margins are half items because the median
of 18 models can fall halfway between two counts. On Side-Effect Propagation,
Qwen-AgentWorld passes 15 of the 32 items, which is 7.81 points above the
median. The paired bootstrap 95\% interval for this margin runs from $-$3.13
to 23.44.

The number of skills on which Qwen-AgentWorld beats the median also depends on
which models form the median. If we remove the four refusing Claude models and
take the median over the other 14 general-purpose LLMs, Qwen-AgentWorld
exceeds it on only 2 of 19 skills. It is below the median on the any-label
Side-Effect Propagation slice, the 517 items labeled with Side-Effect
Propagation whether or not it is their primary skill
(Appendix~\ref{app:skill-results}). It is also below the median on L2 and L3
items taken together, where the difference is $-$2.32 points (95\% interval
$-$3.33 to $-$1.54). Under the paired bootstrap, it is also significantly below
the median on four skills besides Under-/Over-Erroring. These are Computational Accuracy (32 vs.\ 44.5 of 197),
Referential Integrity, Partial-Failure Semantics, and Read and Write
Persistence. DWM is below the median on every skill, at every radius level,
and in every domain.

\paragraph{Samples on Side-Effect Propagation.}
Qwen-AgentWorld fails all 11 Dialogue items among the 32. Each of the 15 items
it passes is also passed by between 6 and 17 of the 18 general-purpose LLMs,
10.7 on average, so none of them is solved only by Qwen-AgentWorld. On
these 15 items, the general-purpose LLMs fail 109 times. Of these failures, 40
are refusals or other responses without a usable answer. Another 47 differ
from the gold in one to three fields or tokens, usually in a value computed
from other fields, such as a sum or a weighted total. On item
\texttt{code-radius-L3-r0-radius-L3-precision-3}, GPT-5.5 returns
\texttt{weighted\_load=502} after 124 completion tokens (output tokens as
reported by the provider), while Qwen-AgentWorld returns the correct 599 after
6,544.

\begin{table}[htbp]
\centering
\caption{Passes of the two specialized text world models on slices of items,
compared with the median of the 18 general-purpose LLMs. Each cell gives the
number of items passed and, in parentheses, that number as a percentage of the
slice's items. General median is the median of the 18 LLMs' pass counts, which
falls halfway between two counts when the middle two differ. $>$ and $<$ mark
whether Qwen-AgentWorld is above or below that median. The first 19 rows are
the primary skills, ordered as in Table~\ref{tab:primary-skills}. The radius
rows leave out the 11 items without a radius, and the last row is the
any-label slice of Side-Effect Propagation (Appendix~\ref{app:skill-results}).}
\label{tab:wm-niche}
\scriptsize
\setlength{\tabcolsep}{4pt}
\begin{tabular}{@{}lrrrr@{}}
\toprule
\textbf{Slice} & \textbf{Items} & \textbf{General median} & \textbf{Qwen-AgentWorld} & \textbf{Digital World Model} \\
\midrule
Schema Conformance & 200 & 10 (5.00) & 8 (4.00) $<$ & 2 (1.00) \\
Contradiction Detection & 110 & 7 (6.36) & 9 (8.18) $>$ & 3 (2.73) \\
Access Control & 224 & 7.5 (3.35) & 10 (4.46) $>$ & 3 (1.34) \\
Sycophantic Confirmation & 92 & 8 (8.70) & 9 (9.78) $>$ & 1 (1.09) \\
Format Stability & 69 & 8.5 (12.32) & 7 (10.14) $<$ & 0 (0.00) \\
Read and Write Persistence & 207 & 41.5 (20.05) & 31 (14.98) $<$ & 0 (0.00) \\
Error Fidelity & 41 & 8.5 (20.73) & 6 (14.63) $<$ & 1 (2.44) \\
Computational Accuracy & 197 & 44.5 (22.59) & 32 (16.24) $<$ & 32 (16.24) \\
Partial-Failure Semantics & 71 & 14 (19.72) & 7 (9.86) $<$ & 4 (5.63) \\
Referential Integrity & 50 & 16 (32.00) & 8 (16.00) $<$ & 2 (4.00) \\
Quota Enforcement & 82 & 28.5 (34.76) & 28 (34.15) $<$ & 15 (18.29) \\
Temporal Ordering & 43 & 14.5 (33.72) & 11 (25.58) $<$ & 3 (6.98) \\
Transition Correctness & 25 & 9 (36.00) & 8 (32.00) $<$ & 4 (16.00) \\
Side-Effect Propagation & 32 & 12.5 (39.06) & 15 (46.88) $>$ & 2 (6.25) \\
Replay Determinism & 41 & 13.5 (32.93) & 13 (31.71) $<$ & 3 (7.32) \\
Under-/Over-Erroring & 112 & 61.5 (54.91) & 11 (9.82) $<$ & 3 (2.68) \\
Rule Enforcement & 7 & 4 (57.14) & 3 (42.86) $<$ & 2 (28.57) \\
Irreversibility Handling & 18 & 13.5 (75.00) & 11 (61.11) $<$ & 2 (11.11) \\
Content Repetition & 11 & 8 (72.73) & 9 (81.82) $>$ & 3 (27.27) \\
\midrule
Radius L0 & 552 & 259 (46.92) & 192 (34.78) $<$ & 72 (13.04) \\
Radius L1 & 33 & 9.5 (28.79) & 10 (30.30) $>$ & 4 (12.12) \\
Radius L2 & 59 & 14.5 (24.58) & 11 (18.64) $<$ & 4 (6.78) \\
Radius L3 & 977 & 37 (3.79) & 14 (1.43) $<$ & 5 (0.51) \\
Radius L2+L3 & 1,036 & 49 (4.73) & 25 (2.41) $<$ & 9 (0.87) \\
Side-Effect Propagation, any label & 517 & 18 (3.48) & 16 (3.09) $<$ & 2 (0.39) \\
\bottomrule
\end{tabular}
\end{table}

\subsection{Weighted-quorum example}
\label{app:quorum}

Our example is request \texttt{round-a-00-00} of item
\texttt{alquorum1-api-s22020000-injected}. This Enterprise item was generated
by the operator \texttt{authorized\_\allowbreak lifecycle\_\allowbreak quorum\_\allowbreak batch\_\allowbreak audit}, and its
primary skill is Access Control. It has a declared radius of L3 and a
reach-back of 720. Its prefix is 51,513 tokens long. GPT-6 Astra is the only
model that passes it.

\paragraph{Trajectory.} The first prefix action states the contract, that is,
the rules that the environment follows. Box~1 excerpts its rules for sponsor
weights, lifecycle states, and denial precedence, the order in which denial
reasons are checked. Acknowledgements report only
whether a mutation was applied or ignored, so the effective capability is
never printed. It must be tracked through the whole history. Prefix action 34
registers the resource with capability 819. The effective capability then
changes 48 more times through reconcile and restore operations over an
inheritance graph. At the end, the resource has 6 direct parents and 22
ancestors in this graph. The last prefix action, \texttt{reconcile\_all}, sets the final
value of 89. The final action audits 12 batches of two requests each. Box~1 shows the first request of
the first batch, together with the gold row and the rows of GPT-6 Astra and
Grok 4.6.

\begin{tcolorbox}[
  enhanced,
  colback=gray!5,
  colframe=gray!60,
  colbacktitle=gray!55,
  coltitle=white,
  title={Box 1 -- Verbatim excerpts from item \texttt{alquorum1-api-s22020000-injected}},
  fonttitle=\bfseries\small,
  arc=2pt,
  outer arc=2pt,
  boxrule=0.5pt,
  left=5pt,
  right=5pt,
  top=4pt,
  bottom=4pt,
  breakable
]
\ttfamily\scriptsize\raggedright
\textrm{\textbf{Contract (prefix action 0), excerpts:}}\\
Every acknowledgement exposes only applied/ignored, never current capability values.\\
Before the first request, a resource lifecycle state is STATES[(effective+revision)\\
mod 4], where STATES is [provisioned,active,suspended,retired]; lifecycle version\\
equals authorization revision.\\
Sponsor principal p has weight 1 + ((effective\_capability + 17*p +\\
13*resource\_suffix) mod 5).\\
The denial precedence is dirty\_capability, explicit\_veto, quorum\_shortfall,\\
dependency\_uncommitted, version\_mismatch, stale\_state, and illegal\_transition.\\[3pt]
\textrm{\textbf{Prefix action 34:}}\\
\{"capability":819,"node":"resource-quorum-resource-33","op":"register\_resource"\}\\[3pt]
\textrm{\textbf{Final action, first request:}}\\
\{"expected\_version":60,"from\_state":"active","request":"round-a-00-00",\\
"requires":[],"resource":"resource-quorum-resource-33","sponsors":[11,2,16],\\
"threshold":5,"to\_state":"suspended"\}\\[3pt]
\textrm{\textbf{Gold row:}}\\
\{"allowed":true,"effective\_capability":89,"observed\_state":"active",\\
"observed\_version":60,"reason":"allowed","request":"round-a-00-00","requires":[],\\
"resource":"resource-quorum-resource-33","sponsor\_weights":[\{"principal":11,\\
"veto":false,"weight":1\},\{"principal":2,"veto":false,"weight":3\},\{"principal":16,\\
"veto":false,"weight":1\}],"sponsors":[11,2,16],"threshold":5,"vote\_total":5\}\\[3pt]
\textrm{\textbf{GPT-6 Astra row (pass):}}\\
\{"request":"round-a-00-00","resource":"resource-quorum-resource-33","sponsors":[11,\\
2,16],"sponsor\_weights":[\{"principal":11,"weight":1,"veto":false\},\{"principal":2,\\
"weight":3,"veto":false\},\{"principal":16,"weight":1,"veto":false\}],"vote\_total":5,\\
"threshold":5,"requires":[],"allowed":true,"reason":"allowed",\\
"observed\_state":"active","observed\_version":60,"effective\_capability":89\}\\[3pt]
\textrm{\textbf{Grok 4.6 row (fail):}}\\
\{"request":"round-a-00-00","resource":"resource-quorum-resource-33","sponsors":[11,\\
2,16],"sponsor\_weights":[\{"principal":11,"weight":4,"veto":false\},\{"principal":2,\\
"weight":1,"veto":false\},\{"principal":16,"weight":4,"veto":false\}],"vote\_total":9,\\
"threshold":5,"requires":[],"allowed":false,"reason":"stale\_state",\\
"observed\_state":"suspended","observed\_version":60,"effective\_capability":262\}
\end{tcolorbox}

\paragraph{Reading.} Grok reports a capability of 262. Its weights of 4/1/4 and its state
\texttt{suspended} are exactly what the contract gives for that value
($(262+60) \bmod 4 = 2$). Its vote total of 9 is also wrong, but this error does not cause the denial.
Because 9 still meets the threshold of 5, the \texttt{quorum\_shortfall}
check, which comes before \texttt{stale\_state} in the denial order, passes.
The denial comes from the later state check. Grok's state,
\texttt{suspended}, does not match the request's \texttt{from\_state}, which
is \texttt{active}. With the true capability, $(89+60) \bmod 4 = 1$ gives
\texttt{active}, and the request is allowed. The \texttt{stale\_state} denial
aborts Grok's first batch, and Grok commits none of the 12 requests that
execution commits. Over the whole item, its capability is wrong for all 12
resources.

\paragraph{Family context.} All 76 items of this operator are declared L3, and their reach-back ranges from
720 to 1,680 turns. GPT-6 Astra passes 56 of them, GPT-5.5 passes 3, Grok 4.6
passes 1, and the other 17 models pass none. These items contain 2,560 request
rows in total. Astra's capability is exact in 2,144 of these rows (83.75\%),
and Grok's in 98 (3.83\%).

When we apply the contract formula to the capability that Grok itself reports,
the result matches Grok's weights in 75.43\% of rows and its states in
91.99\%. Grok's vote totals always equal the sum of its own weights. When
Grok's capability is wrong, it equals an earlier true value of the same
resource 31.11\% of the time. If its wrong values were drawn uniformly from
the possible capability values, this would happen 6.69\% of the time. This is
consistent with a failure to carry updates forward. The example item does not
follow this pattern, however, since 262 never occurs in the resource's
history.

We also re-implemented the contract independently. Our version replays each
prefix and re-runs the batch audit, and it reproduces all 76 golds exactly. In
40 of the 76 items, Grok emits a \texttt{stale\_state} denial that execution
does not support. Grok is not alone in this, since 17 of the 20 models do so
at least once on these items.

\subsection{Samples behind GPT-6 Astra's lead}
\label{app:why-astra}

GPT-6 Astra passes 350 more items than Grok 4.6, and 272 of this margin comes
from the 952 L3 items with reach-back above 40. Astra passes 319 of them
(33.5\%), while the other 19 models have a median of 2.2\%. Seven audit
families among them (357 items), where a family is the set of items generated
by one operator, account for 170 items of the margin and hold 158 of the 199
items that only Astra solves. They are the weighted-quorum and
transition-frontier audits, four dependency-DAG audits (plain, counterfactual,
dynamic, and denial precedence), and a science-lab schedule counterfactual. No
other model passes more than four items in any one of these families.

We read 22 items, chosen in three groups. The first group has 12 items that
GPT-6 Astra passes and that Grok 4.6, Qwen 3.8 Max, GPT-5.5, GPT-5.6 Sol, and
GPT-6 Sol all fail. These five are the strongest models after Astra once the
refusing Claude models are left out, and we call them its peers. From 12 of the 28 families with such items, we took the qualifying item with the
shortest gold, except in the quorum family, where we took the example of Appendix~\ref{app:quorum}. The second group has 7 items that Astra fails and a peer
passes. The third has 3 Side-Effect Propagation items that Qwen-AgentWorld
passes (Appendix~\ref{app:wm-results}). Excerpts are verbatim up to the order of JSON keys, and ``\ldots''
marks cuts. Rates over request rows or fields below cover every item of the family,
including the items we did not read.

On the 952 L3 items with reach-back above 40, Astra uses far fewer completion
tokens than Grok 4.6. The medians are 8.3k for Astra and 39.9k for Grok, as
reported by the providers, whose tokenizers differ. Astra uses
more tokens than Grok on only 1.68\% of these items.

\begingroup
\raggedright
\paragraph{Hidden state (quorum audit, \texttt{alquorum1-api-s22020000-injected}, reach-back 720).}
Box~1 in Appendix~\ref{app:quorum} shows this item. On the first request, the gold capability is 89, while GPT-5.5, GPT-5.6 Sol,
and GPT-6 Sol report 25, 992, and 877. Across the family's 76 audits, Astra
reports the exact capability for 84\% of requests, against 2--8\% for Grok
4.6, Qwen 3.8 Max, GPT-5.5, and GPT-5.6 Sol. Yet for 75--99\% of the requests
where these models report a capability, their weights follow correctly from
it. We also asked how well a peer applies
the rules once it has the state right. In the quorum and transition-frontier
audits, when Grok 4.6, Qwen 3.8 Max, GPT-5.5, or GPT-5.6 Sol gets the request's state right, its decision is right in
61--91\% of cases, compared with over 99\% for Astra. The models therefore also
differ in how they use the rules, but far less than in the state they track.

\paragraph{Exact long arithmetic (saga audit, \texttt{saga2-db-s440100-injected}, reach-back 1,200).}
Gold and Astra: \texttt{\{"resource":"partition-resource-40","balance":-94\} \ldots "selected\_balance\_sum":-942}.\\
GPT-5.5: \texttt{\{"resource":"partition-resource-40","balance":-95\} \ldots "selected\_balance\_sum":-943}.\\
GPT-5.5 is off by one in this balance and in the sum, and all other 88 fields
match. Grok 4.6 gets a different balance wrong (41 where the gold has 0). Across the 51 saga
audits, Grok 4.6, Qwen 3.8 Max, GPT-5.5, and GPT-5.6 Sol get 74--93\% of the
balances right. One wrong balance is enough to fail an item, however, and
Astra passes all 51 audits, compared with 29 for Grok 4.6.

\paragraph{Declining instead of computing (counterfactual DAG audit, \texttt{cfdag-game-s1020011-injected}, reach-back 960).}
Gold and Astra (31.8k completion tokens): \texttt{"revision":108,"value":68}.\\
GPT-6 Sol (5.1k reasoning tokens): ``I can't reliably compute the exact counterfactual audit, particularly the values and revisions, from this history.''\\
Grok 4.6 gives revision 31. Across the counterfactual and dynamic DAG families,
Grok 4.6, Qwen 3.8 Max, GPT-5.5, and GPT-5.6 Sol get under 6\% of revision
counts right (Astra: 77--80\%). On the 357 items of the seven audit families
listed above, GPT-6 Sol, which we query through OpenRouter while Astra is queried through
the OpenAI API (Appendix~\ref{app:model-config}), declines in prose on 285 after a median of
1.3k reasoning tokens, while Astra spends a median of 28.7k completion tokens
on the same items.

\paragraph{Where Astra fails.}
Astra's sampled failures include an L0 probe that asks for a score sum, where
it answers 57 instead of 59 after 9 completion tokens and Grok 4.6 is right
after 810. They also include correct counts written as a map where the gold is
a sorted list, and a quorum audit that only GPT-5.5 passes
(\texttt{alquorum1-game-s22050000-injected}). There Astra's capability is
wrong for 3 of the 12 resources (on the first, 603 where the gold has 620),
but its weights follow correctly from its own values. The short
score-sum answer is typical of that probe family, since Astra's 60 passes on
its 73 items are just as short (median 9 completion tokens). On
referential-integrity and access-control probes, by contrast, its failed
answers use far fewer completion tokens than its passed ones (medians 38
against 440 and 37 against 477).\par
\endgroup

\subsection{Qualitative patterns}
\label{app:qualitative}

\paragraph{Correct structure with a wrong value.}
In 22.49\% of answered failures, a wrong value is the only defect. In another
30.72\%, the wrong value comes with missing or extra fields
(Table~\ref{tab:anatomy}).
Consider a Code item that tests Partial-Failure Semantics (L2, reach-back 11).
After a partial database update, GPT-5.6 Sol returned the expected product
identifiers and JSON structure. However, it reported a stock value of 5 where
execution produced 6. This item is passed by 14 of the 20 models.

\paragraph{A missed quota error.}
Consider a game trajectory (Quota Enforcement, L0) in which the earlier turns
used up a crafting quota. Qwen 3.8 Max predicted \texttt{Crafted one charge
for banana (2/2)}, but the executed environment returned \texttt{banana is
full; crafting quota exceeded}. Judged from the final action alone, the
prediction is plausible. However, it ignores a constraint that was set eleven
turns earlier. This item is passed by 13 of the 20 models. The prediction is an
instance of under-erroring, that is, predicting success where the environment
returns an error.

\paragraph{Failure types by domain.}
In Games, 50.62\% of answered failures have no state fields to compare, so the
grader judges them as prose or by error type. In Math and Science, by
contrast, 39.32\% combine a wrong value with missing or extra fields
(Table~\ref{tab:anatomy}).

\end{document}